%% file: main.tex
\documentclass[sigconf]{acmart}

\usepackage{subfigure}
\usepackage{pdfpages}
\usepackage{multicol}
\usepackage{graphicx}
\usepackage{bm}
\usepackage{color}
\usepackage{multirow}
\usepackage{enumitem}
\usepackage{algpseudocode}
\usepackage{amsmath,amsfonts,bm}
\usepackage[ruled,linesnumbered]{algorithm2e}

\definecolor{brown}{RGB}{139,64,0}
\definecolor{pink}{RGB}{255,170,182}
\definecolor{purple}{RGB}{160,32,240}

\AtBeginDocument{%
  }

\setcopyright{acmlicensed}
\copyrightyear{2018}
\acmYear{2018}
\acmDOI{XXXXXXX.XXXXXXX}

\acmConference[XXX]{XXX}{June 03--05,
  2018}{Woodstock, NY}
\acmISBN{978-1-4503-XXXX-X/18/06}

\begin{document}



\title{Geometric Flow Matching for Molecular Conformation Generation via Manifold Decomposition}

\author{Yunqing Liu}
\affiliation{%
  \institution{The Hong Kong Polytechnic University}
  \city{Hong Kong}
  \country{China}
}
\email{yunqing617.liu@connect.polyu.hk}

\author{Yi Zhou}
\affiliation{%
  \institution{The Hong Kong Polytechnic University}
  \city{Hong Kong}
  \country{China}
}
\email{echo-yi.zhou@connect.polyu.hk}

\author{Wenqi Fan}
\affiliation{%
  \institution{The Hong Kong Polytechnic University}
  \city{Hong Kong}
  \country{China}
}
\email{wenqifan03@gmail.com}

\renewcommand{\shortauthors}{Liu et al.}

\input{sections/abs}





\maketitle

\input{sections/intro}
\input{sections/related}

\input{sections/methods}

\input{sections/experiments}


\input{sections/conclusion}


\balance
\bibliographystyle{ACM-Reference-Format}
\bibliography{example_paper}


\appendix

\input{sections/appendix}

\end{document}

%% file: sections/abs.tex
\begin{abstract}

The generation of accurate 3D molecular conformations is a pivotal challenge in computational chemistry and drug discovery. Recently, diffusion and flow matching models have achieved remarkable success. However, there is a critical misalignment between their mathematical formulation and the physical reality of molecules. Existing approaches predominantly treat molecules as unstructured point clouds in Cartesian space, overlooking the intrinsic hierarchical mechanics where bond lengths and bond angles are relatively stiff, whereas torsion angles constitute the dominant flexible degrees of freedom. This lack of manifold awareness forces models to relearn fundamental geometric constraints from scratch, often leading to physically implausible intermediate structures. To address this, we propose GO-Flow that aligns generative modeling with molecular geometry via manifold decomposition. Instead of forcing motion through Euclidean space, GO-Flow decomposes the generation process into three physically motivated subspaces: translation space with linear optimal transport, rotation space with geodesic flows on $SO(3)$, and conformation space with entropic optimal transport. This decomposition injects geometric inductive biases and makes the generative paths better aligned with molecular degrees of freedom. When combined with equivariant neural architectures, it encourages rotation-consistent generation and improves geometric validity. Extensive experiments on GEOM-Drugs and GEOM-QM9 demonstrate that GO-Flow achieves state-of-the-art generation quality. Notably, by learning naturally straighter probability paths on the correct manifolds, our method enables high-fidelity sampling with as few as 50 steps, effectively bridging the gap between structural precision and computational efficiency. 

\end{abstract}

%% file: sections/intro.tex
\section{Introduction}

Predicting valid spatial coordinates from a 2D molecular graph to 3D molecular conformation is a pivotal challenge in computational chemistry and drug discovery~\citep{li2024speak,li2024tomg,liu-etal-2025-glprotein,li2026molreflect}. Accurate 3D structures determine physicochemical properties and drug-target interactions, acting as the prerequisite for virtual screening and docking~\citep{liu2026enhancing,ding2019selective,weng2021late}. While traditional sampling methods, such as molecular dynamics, are computationally expensive~\cite{hollingsworth2018molecular,singh2022molecular,li2024empowering}, deep generative models have recently emerged as a powerful paradigm~\cite{hu2023improving,zhao2024recommender}. In particular, diffusion probabilistic models (DPMs)~\cite{liu2023generative,ho2020denoising,liu2025score,he2025graph} and flow matching frameworks~\cite{lipman2023flow} have achieved remarkable success by modeling the generation process as a progressive transformation from a prior noise distribution to the data distribution, effectively capturing the multimodal nature of molecular geometries~\cite{sadybekov2023computational,dara2022machine,vincent2022phenotypic}.

\begin{figure}[t]
    \centering
    \includegraphics[width=1\linewidth]{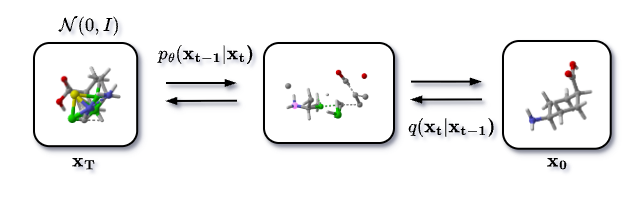}
    \captionsetup{font=small, skip=0pt}
  \vskip -0.1in
    \caption{Illustration of the geometric inconsistency in Cartesian-based molecular generation. Existing methods typically treat molecules as unstructured point clouds in Euclidean space ($\mathbb{R}^{3N}$), applying Gaussian noise uniformly to all atomic coordinates.}
    \Description{Comparison between Cartesian-based molecular generation and geometry-aware molecular generation.}
        \vskip -0.2in
    \label{fig:mol_diff}
\end{figure}
However, a critical misalignment exists between the mathematical formulation of current generative models and the physical reality of molecules. Prevailing approaches predominantly treat molecules as unstructured point clouds in Cartesian Euclidean space ($\mathbb{R}^{3N}$)~\citep{luo2021predicting,shi2021learning}. While mathematically convenient, this assumption ignores the intrinsic hierarchical nature of molecular mechanics~\cite{liu2026enhancing}: molecules are not amorphous clouds but structured entities governed by rigid bond lengths and flexible torsion angles.
Operating blindly in Euclidean space introduces severe geometric inconsistencies~\cite{pmlr-v235-geist24a}. For instance, treating rotational orientation as linear displacement in $\mathbb{R}^3$ violates the geometry of the rotation group $SO(3)$, leading to distortions that must be corrected during training. Similarly, applying uniform Gaussian noise to all atoms forces the model to relearn fundamental chemical constraints, such as bond rigidity, from scratch, as shown in Figure \ref{fig:mol_diff}. This lack of manifold awareness limits the model's ability to capture the true underlying Boltzmann distribution effectively, often resulting in chemically invalid intermediate structures or mode collapse.

To tackle the aforementioned challenges, we propose GO-Flow, a manifold-aware molecular conformation generation framework via \underline{\textbf{G}}eodesic and \underline{\textbf{O}}ptimal transport \underline{\textbf{Flow}}s. 
Instead of forcing molecular motion into a monolithic Cartesian coordinate system, we decompose the generation into three physically motivated subspaces, each modeled with a flow matching objective tailored to its specific geometry.
Specifically, we model the center of mass movement using linear optimal transport, handle global orientation via geodesic flows on the $SO(3)$ manifold (using quaternion representation and spherical linear interpolation), and navigate the internal conformation space using entropic optimal transport. By explicitly modeling these geometric degrees of freedom, GO-Flow provides a manifold-aware parameterization that is compatible with rotation-equivariant architectures and biases the generation process toward chemically plausible conformations. This decomposition allows us to inject physical inductive biases directly into the generative process, bridging the gap between physical interpretability and efficiency.

In summary, our main contributions are highlighted as follows:
\begin{itemize}[leftmargin=*]
    \item We propose GO-Flow, a novel paradigm that departs from unstructured Cartesian denoising by decomposing molecular generation into three physically motivated manifolds: translation ($\mathcal{M}_T$), rotation ($\mathcal{M}_R$), and conformation ($\mathcal{M}_C$). This decomposition allows us to inject physical inductive biases directly into the generative process, effectively bridging the gap between structural precision and physical interpretability.
    \item We introduce tailored flow matching objectives that strictly adhere to the geometry of each subspace. Specifically, we employ geodesic interpolation on the rotation manifold to avoid singular parameterizations and to maintain consistency with rotation-equivariant modeling, and entropic optimal transport for internal coordinates to navigate the complex multimodal energy landscape of bond lengths and torsion angles. This rigorous treatment eliminates the need for the model to relearn basic geometric constraints from scratch.
    \item Extensive experiments on the GEOM-Drugs and GEOM-QM9 benchmarks demonstrate that GO-Flow achieves state-of-the-art performance. By modeling motion on the correct manifolds, our method produces high-fidelity structures with superior geometric validity and diversity compared to Cartesian-based diffusion and flow matching baselines.
\end{itemize}

%% file: sections/related.tex
\section{Related Work}

Molecular conformer generation has long been a fundamental task in computational chemistry, aiming to generate the set of low-energy 3D structures, i.e., conformers, conditioned on the molecular graph~\cite{fan2025computational}. Traditionally, cheminformatics methods are implemented in software such as RDKit~\cite{landrum2013rdkit} and OMEGA~\cite{hawkins2010conformer}. In recent years, a number of machine learning methods have been developed and are highly regarded for their advantages in accuracy. 

Early machine learning methods struggle to address the roto-translation equivariance of conformers~\cite{mansimov2019molecular}. Therefore, a series of VAE or diffusion models is developed based on the distance geometry, learning the distance between atoms rather than the atomic coordinates, representatively the GraphDG~\cite{simm2019generative}, CGCF~\cite{xu2021learning}, ConfVAE~\cite{xu2021end}, and 
ConfGF~\citep{shi2021learning}. 
However, the back-and-forth conversion between the Cartesian coordinate system and the intermediate distance geometric elements resulted in accumulated errors. 
Models might learn from invalid distance matrices and produce incorrect structures. 
Subsequently, GeoDiff~\citep{xu2022geodiff} makes a big step by theoretically demonstrating that Markov chains evolving with equivariant Markov kernels can induce an invariant distribution by design, therefore enabling direct end-to-end acts on the atomic Cartesian coordinate system. 
The diffusion/flow matching methods in Euclidean space have gradually become mainstream.
EBD~\cite{park2024equivariant} designs two hierarchical stages to generate atomic details from fragment-level coarse-grained
structures.
MCF~\cite{wang2024swallowing} shows that simply scaling up model capacity yields large gains in generalization performance.
AvgFlow~\cite{cao2025efficient} is built upon flow-matching and proposes SO(3)-averaged flow and reflow/distillation strategies for accelerated training and inference.

Nevertheless, another branch of studies claims that leveraging appropriate geometric prior knowledge (in addition to distance geometry) offers advantages for learning geometric consistency and improving sampling efficiency. 
GeoMol~\citep{ganea2021geomol} directly predicts a minimal set of geometric quantities (torsion angles, local bond distance, and local bond angles) sufficient for deterministic reconstruction of the 3D conformer, in a highly efficient single forward manner. 
TorDiff~\cite{jing2022torsional} operates on the space of torsion angles via a diffusion process, using two orders of magnitude fewer denoising steps than GeoDiff, the representative Euclidean diffusion approaches.
ET-Flow~\cite{hassan2024flow} is a well-designed flow-matching approach that operates directly on all-atom molecular coordinates with minimal equivariance and harmonic prior, requiring significantly fewer parameters and sampling steps.
They demonstrate the promising potential of generative molecular modeling with manifold awareness in addition to the Cartesian-centric manner.

%% file: sections/methods.tex
\section{Methodology}

\begin{figure*}[htp]
    \centering
    \includegraphics[width=1.1\linewidth]{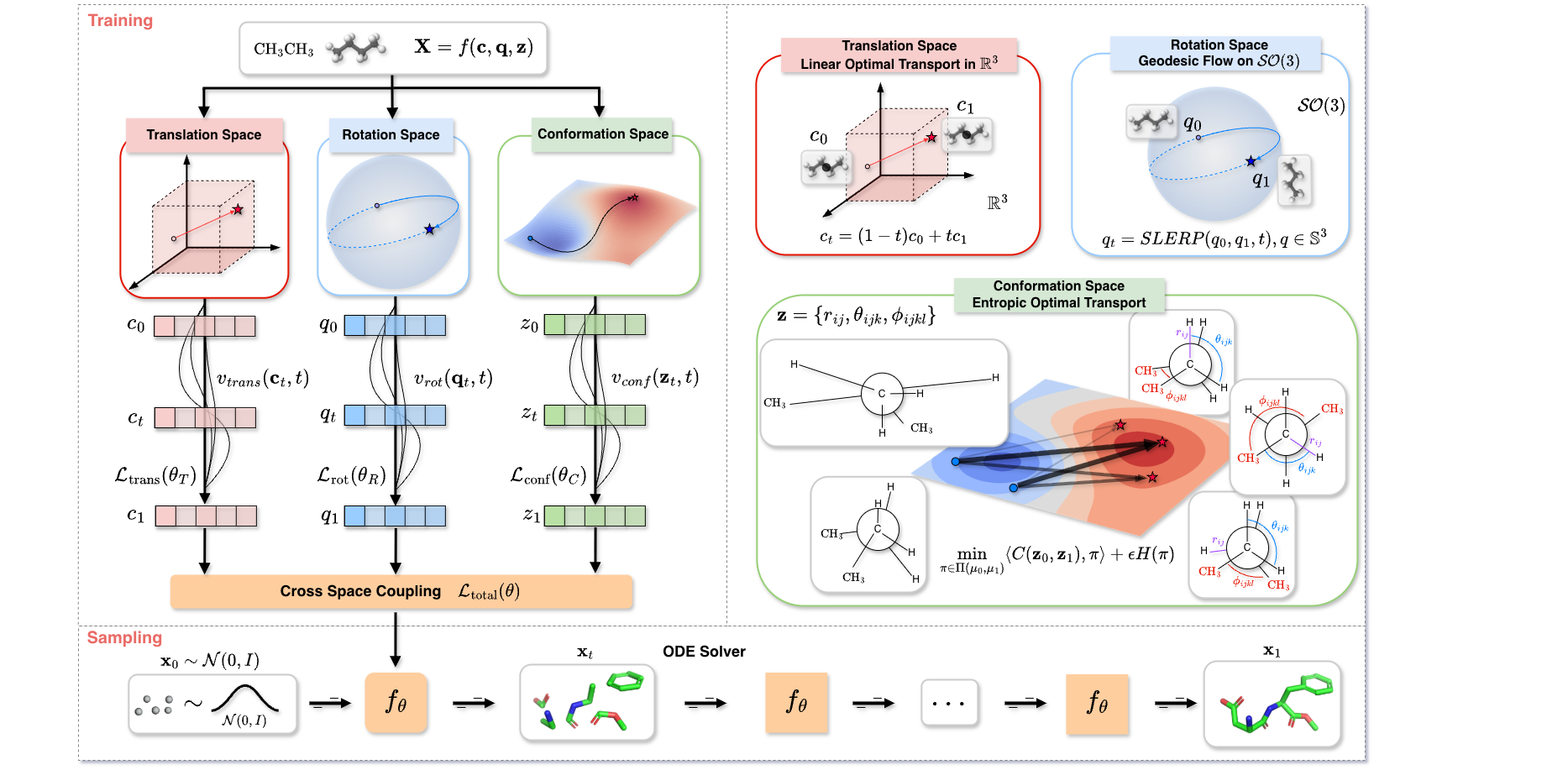}
    \captionsetup{font=small, skip=3pt}
    \caption{An illustration of the proposed GO-Flow framework. The framework is divided into Training (top) and Sampling (bottom) phases. \emph{Training}: The model decomposes molecular generation into three physically motivated subspaces to learn distinct velocity fields: Translation Space, Rotation Space and Conformation Space. \emph{Sampling}: During inference, the model starts from a Gaussian noise distribution $x_0 \sim \mathcal{N}(0, I)$ and employs an ODE Solver to integrate the learned vector fields $f_\theta$, generating the final physically valid 3D molecular conformation $x_1$.}
            \vskip -0.12in

    \label{fig:overview}
\end{figure*}

As shown in Figure \ref{fig:overview}, we present GO-Flow, a geometric flow matching framework that addresses the challenge of 3D molecular generation by decomposing the generative process into three physically motivated manifolds: translation ($\mathcal{M}_T = \mathbb{R}^3$), rotation ($\mathcal{M}_R = SO(3)$), and internal conformation ($\mathcal{M}_C = \mathbb{R}_+^{n_r} \times (0,\pi)^{n_\theta} \times \mathbb{T}^{n_\phi}$), where bond lengths, bond angles, and torsion angles are modeled with their corresponding domains. Unlike Cartesian-based diffusion, which treats molecules as point clouds, GO-Flow learns velocity fields on these specific manifolds to ensure geometric consistency and sampling efficiency.

\subsection{Problem Formulation}

Given a molecular graph $\mathcal{G} = (\mathcal{V}, \mathcal{E})$ with atoms $\mathcal{V}$ and bonds $\mathcal{E}$, our goal is to generate 3D conformations $\mathbf{X} \in \mathbb{R}^{N \times 3}$ that are chemically valid and energetically favorable. Existing diffusion models predominantly treat molecules as point clouds in Euclidean space, applying noise uniformly to all Cartesian coordinates. This assumption is physically flawed: it overlooks the intrinsic hierarchy of molecular mechanics, where bond lengths are chemically rigid while torsion angles are flexible. Consequently, Cartesian-based methods must "re-learn" basic geometric constraints (e.g., bond lengths) from scratch, resulting in slow convergence and physically invalid intermediate structures. To bridge the gap between geometric generation and physical interpretability, GO-Flow decomposes the molecular state space into three physically motivated manifolds, as illustrated in Figure ~\ref{fig:overview}:
\begin{equation}
    \mathbf{X} = f(\mathbf{c}, \mathbf{q}, \mathbf{z}),
\end{equation}
where $c\in \mathbb{R}^3$ is the center of mass, $q\in S^3/\{\pm 1\}$ represents a rotation in $SO(3)$, and $z = (r,\theta,\phi) \in \mathbb{R}_+^{n_r} \times (0,\pi)^{n_\theta} \times \mathbb{T}^{n_\phi}$ denotes internal coordinates consisting of bond lengths. bond angles, and periodic torsion angles.
We define the generative process not as a single diffusion trajectory, but as three coupled flows on distinct manifolds: translation ($\mathcal{M}_T$), rotation ($\mathcal{M}_R$), and conformation ($\mathcal{M}_C$).

\subsection{Manifold-Aware Flow Matching}
We construct probability paths maximizing the likelihood of data by learning velocity fields $v_t$ that transport a prior distribution (noise) to the data distribution.

\subsubsection{Translation Space}

Global position is chemically irrelevant but statistically confounding if mixed with internal structure. By explicitly isolating the center of mass, we reduce the complexity of the learning task for the remaining subspaces.

We model the center of mass $c \in \mathbb{R}^3$ using linear optimal transport. The interpolation path between the noise distribution $c_0$ and target $c_1$ is linear, yielding a constant velocity field that simplifies to a regression objective:
\begin{align}
    &v_{trans}(c_t, t) = c_1 - c_0   ,\\
    &c_t = (1-t)c_0 + t c_1.
\end{align}
This ensures the simplest possible transport path for global displacement.

\subsubsection{Rotation Space}
Treating molecular orientation as linear displacements in $\mathbb{R}^3$ ignores the intrinsic curvature of the rotation manifold. Linear interpolation of rotation matrices or Euler angles introduces singularities (gimbal lock) and geometric distortions (shrinking norms)~\cite{hemingway2018perspectives}. To guarantee strictly rotation-equivariant generation, the generative flow must evolve along the geodesics of the special orthogonal group $SO(3)$.

We represent rotations by unit quaternions $q \in \mathbb{S}^3$, with the antipodal identification $q \sim -q$, since unit quaternions form a double cover of $SO(3)$. The optimal path between a random orientation $q_0$ and the target $q_1$ is the shortest arc on the hypersphere, computed via spherical linear interpolation (SLERP)~\cite{buss2001spherical}:
\begin{align}
    q_t = \text{SLERP}(q_0, q_1, t) =\frac{\sin((1-t)\Omega)}{\sin \Omega}q_0 + \frac{\sin(t\Omega)}{\sin \Omega}q_1,
\end{align}
where $\Omega = \arccos(|q_0 \cdot q_1|)$.

To train the flow matching model, we must derive the tangent space velocity (angular velocity $\omega$) rather than the quaternion derivative. Differentiating the SLERP path (see Appendix \ref{app:A}), the target rotational velocity field is:
\begin{align}
    &v_{rot}(q_t, t) = \frac{1}{2}\omega q_t, \\ 
    &\omega = 2 \cdot \text{Im}(\log(q_1 q_0^*)). 
\end{align}
Here, $\log(\cdot)$ maps the relative rotation to the Lie algebra $SO(3)$, represented as an angular velocity vector, ensuring the velocity physically corresponds to a valid 3D rotation axis and magnitude.

\subsubsection{Conformation Space}

This manifold encapsulates the "chemistry" of the molecule. The space of internal coordinates $z$ (bond lengths, angles, torsions) is highly multimodal with a complex energy landscape. Standard Gaussian paths often cross high-energy barriers or generate sterically clashing structures. We require a transport plan that respects the distribution's geometry, necessitating entropic optimal transport rather than simple linear interpolation.

We formulate the transport as finding the optimal coupling $\pi^*$ that minimizes the regularized transport cost:
\begin{align}
    \min_{\pi \in \Pi(\mu_0, \mu_1)} \langle C(z_0, z_1), \pi \rangle + \epsilon H(\pi),
\end{align}
where $C(z_0, z_1) = \|z_0 - z_1\|_2^2$ and $H(\pi)$ is the entropy. By using the Sinkhorn algorithm, we compute $\pi^*$ efficiently. The target velocity field $v_{conf}$ is defined by the barycentric projection of this optimal plan:
\begin{align}
    v_{conf}(z_t, t) = \mathbb{E}_{\pi^*}[z_1 - z_0 | z_t].
\end{align}
This entropic regularization effectively smooths the vector field, preventing overfitting to specific conformers and allowing the model to capture the diversity of the Boltzmann distribution.

\subsection{Geometric Composition via Jacobians}

While we learn flows on decomposed manifolds, the final output must be Cartesian coordinates for downstream applications. A naive combination would violate the geometric constraints we worked hard to preserve. We therefore employ a mathematically rigorous Jacobian-based projection to map internal velocities back to Cartesian updates. The total atomic velocity $v_{total}$ is the composition of the three subspaces:
\begin{align}
    v_{total} = v_{trans}^{cart} + \underbrace{\omega \times (r - c_m)}_{v_{rot}^{cart}} + \underbrace{J(z) v_{conf}}_{v_{conf}^{cart}},
\end{align}
where $J(z) \in \mathbb{R}^{3N \times m}$ is Jacobian matrix, including bond length Jacobian, bond angle Jacobian and dihedral angle Jacobian. We explicitly derive its components to ensure stable backpropagation (Appendix \ref{app:derivations}).

\begin{table*}[htp]
  \caption{Geometric evaluation on GEOM-Drugs ($\delta = 1.25\text{\r{A}}$).}
  \label{tab:rmsd_drug}
  \centering
  \vspace{-5pt}
  \resizebox{0.8\textwidth}{!}{
    \begin{tabular}{l|c|cccc|cccc}
    \toprule[1.0pt]
    & & \multicolumn{2}{c}{\shortstack[c]{COV-R ($\%$) $\uparrow$}}  & \multicolumn{2}{c|}{\shortstack[c]{MAT-R($\text{\r{A}}$) $\downarrow$}}  & \multicolumn{2}{c}{\shortstack[c]{COV-P ($\%$) $\uparrow$}}  & \multicolumn{2}{c}{\shortstack[c]{MAT-P ($\text{\r{A}}$) $\downarrow$}} \\
    Models & Steps & Mean & Med & Mean & Med & Mean & Med & Mean & Med \\
    \midrule[0.8pt]
    RDKit & - & 45.74 & 31.75 & 1.5376 & 1.4004 & 54.78 & 59.48 & 1.3341 & 1.1996 \\ 
    GraphDG & - & 8.27 & 0.00 & 1.9722 & 1.9845 & 2.08 & 0.00 & 2.4340 & 2.4100 \\ 
    CGCF & - & 53.96 & 57.06 & 1.2487 & 1.2247 & 21.68 & 13.72 & 1.8571 & 1.8066 \\
    ConfVAE& -  & 55.20 & 59.43 & 1.2380 & 1.1417 & 22.96 & 14.05 & 1.8287 & 1.8159 \\
    GeoMol& -  & 67.16 & 71.71 & 1.0875 & 1.0586 & - & - & - & - \\ 
    ConfGF& -  & 62.15 & 70.93 & 1.1629 & 1.1596 & 23.42 & 15.52 & 1.7219 & 1.6863 \\
    \midrule
    GeoDiff& 5,000 & 89.40 & 96.86 & 0.8571 & 0.8495 & 61.28 & 65.00 & 1.1642 & 1.1272 \\
    MCF & 50 & 76.47 & 79.21 & 1.0709 & 0.9952 & 56.96 & 61.40 & 1.5937 & 1.4629 \\
    EBD & 50 & 89.37 & 93.21 & 0.8216 & 0.8279 & 66.24 & 68.39 & 1.1237 & 1.0916 \\ 
    AvgFlow & 50 & 91.72 & 95.70 & 0.8044 & 0.7978 & 68.56 & 70.10 & 1.1294 & 1.1131 \\
    ET-Flow & 50 & 93.12 & 99.03 & 0.8049 & 0.8004 & 69.41 & 69.90 & 1.1244 & 1.1037 \\
    \textbf{GO-Flow} & 50 & \textbf{94.82}& \textbf{99.26} &\textbf{0.7971} & \textbf{0.7924} & \textbf{70.13} & \textbf{72.49} & \textbf{1.1068} & \textbf{1.0001} \\ 
    \bottomrule[1.0pt]
    \end{tabular}
    }
\end{table*}

\subsection{Training Strategy}
\begin{algorithm}
\caption{GO-Flow Three-Stage Training}
\label{alg:training}

\textbf{Stage 1: Disentangled Manifold Learning} ($E_1$ epochs) \\

\For{$s \in \{\mathrm{trans},\mathrm{rot},\mathrm{conf}\}$}{
    \For{$e=1$ to $E_1$}{
        \For{batch $\mathbf{x}_1 \sim \mathcal{D}$}{
            Decompose $(\mathbf{c}_1,\mathbf{q}_1,\mathbf{z}_1)=\mathrm{decompose}(\mathbf{x}_1)$ \\
            Sample $(\mathbf{c}_0,\mathbf{q}_0,\mathbf{z}_0)\sim p_0$ and $t\sim U(0,1)$ \\

            $\mathbf{c}_t=(1-t)\mathbf{c}_0+t\mathbf{c}_1$ \\

            $\widetilde{\mathbf{q}}_1=\mathbf{q}_1$ if $\mathbf{q}_0^\top\mathbf{q}_1\ge0$; otherwise $\widetilde{\mathbf{q}}_1=-\mathbf{q}_1$ \\
            $\mathbf{q}_t=\mathrm{SLERP}(\mathbf{q}_0,\widetilde{\mathbf{q}}_1,t)$ \\

            Compute entropic OT coupling $\pi^*$ for internal coordinates \\
            Pair $(\mathbf{z}_0,\mathbf{z}_1)\sim \pi^*$ and construct $\mathbf{z}_t=\mathrm{Interp}_{\mathcal{M}_C}(\mathbf{z}_0,\mathbf{z}_1,t)$ \\

            Compute the space-specific target velocity $u_s^*$ \\
            $\mathcal{L}_s=\|v_s(\cdot_t,t)-u_s^*\|_2^2$ \\

            $\theta_s\leftarrow\theta_s-\alpha_s\nabla_{\theta_s}\mathcal{L}_s$ \\
        }
    }
}

\textbf{Stage 2: Cross-Space Coupling} ($E_2$ epochs) \\

Initialize $\theta=\{\theta_T^{(E_1)},\theta_R^{(E_1)},\theta_C^{(E_1)}\}$ \\

\For{$e=1$ to $E_2$}{
    \For{batch $\mathbf{x}_1\sim\mathcal{D}$}{
        Construct $\mathbf{c}_t,\mathbf{q}_t,\mathbf{z}_t$ as in Stage 1 \\
        Compose $\mathbf{x}_t=g(\mathbf{c}_t,\mathbf{q}_t,\mathbf{z}_t)$ \\

        Compute decomposed velocities $v_T,v_R,v_C$ \\
        Compose Cartesian velocity:
        \[
        v_{\mathrm{total}}
        =
        v_T^{\mathrm{cart}}
        +
        \boldsymbol{\omega}_{\theta}\times(\mathbf{r}_t-\mathbf{c}_t)
        +
        J(\mathbf{z}_t)v_C
        \]
        Compute target Cartesian velocity $\dot{\mathbf{x}}_t^*$ from the manifold path \\

        $\mathcal{L}_{\mathrm{flow}}
        =
        \|v_{\mathrm{total}}(\mathbf{x}_t,t)-\dot{\mathbf{x}}_t^*\|_2^2$ \\

        $\lambda_s\leftarrow \frac{1}{2\sigma_s^2}$ \\

        $\mathcal{L}_{\mathrm{total}}
        =
        \sum_{s\in\{T,R,C\}}\lambda_s\mathcal{L}_s
        +
        \lambda_F\mathcal{L}_{\mathrm{flow}}$ \\

        $\theta\leftarrow\theta-\alpha\nabla_\theta\mathcal{L}_{\mathrm{total}}$ \\
    }
}

\textbf{Stage 3: ODE Fine-tuning} ($E_3$ epochs) \\

\For{$e=1$ to $E_3$}{
    \For{batch $\mathbf{x}_1\sim\mathcal{D}$}{
        Integrate the CNF ODE backward:
        \[
        \frac{d\mathbf{x}_t}{dt}=v_\theta(\mathbf{x}_t,t),
        \qquad
        \frac{d\log p_t(\mathbf{x}_t)}{dt}
        =
        -\mathrm{Tr}
        \left(
        \frac{\partial v_\theta}{\partial \mathbf{x}_t}
        \right)
        \]
        Estimate the trace by Hutchinson estimator \\
        Compute $\log p_1(\mathbf{x}_1)$ and update:
        \[
        \theta\leftarrow
        \theta-\alpha\nabla_\theta[-\log p_1(\mathbf{x}_1)]
        \]
    }
}

\end{algorithm}

The training objective combines flow matching losses from all
three coordinate spaces. We derive each loss component from the
corresponding optimal transport formulation. For linear optimal transport in translation space, the flow matching loss is
\begin{equation}
    \mathcal{L}_{\text{trans}} = \mathbb{E}_{t,\mathbf{c}_0,\mathbf{c}_1} \left[ \|v_{\text{trans}}(\mathbf{c}_t, t) - (\mathbf{c}_1 - \mathbf{c}_0)\|^2 \right] ,
\end{equation}
where $\mathbf{c}_t = (1-t)\mathbf{c}_0 + t\mathbf{c}_1$ and the target velocity is the constant vector $\mathbf{c}_1 - \mathbf{c}_0$.
For the rotation component, we construct the target path using geodesics on $SO(3)$, and train the network by matching the corresponding angular velocity in the Lie algebra. Since angular velocities lie in Lie algebra, we use an Euclidean loss in $SO(3)$
\begin{equation}
    \mathcal{L}_{\text{rot}} = \mathbb{E}_{t,\mathbf{q}_0,\mathbf{q}_1} \left[ ||\hat{\boldsymbol{\omega}}_\theta (\mathbf{q}_t,t) - \boldsymbol{\omega}_{target}||^{2}_{2}
    \right].
\end{equation}

For optimal transport in internal coordinate space, we use the Wasserstein distance
\begin{equation}
    \mathcal{L}_{\text{conf}} = \mathbb{E}_{t,\mathbf{z}_0,\mathbf{z}_1} \left[ W_2^2(v_{\text{conf}}(\mathbf{z}_t, t), v_{\text{OT}}(\mathbf{z}_t, t)) \right], 
\end{equation}
where $v_{\text{OT}}(\mathbf{z}_t, t)$ is the velocity field induced by the optimal transport plan $\pi^*$. 
The velocity field is computed as 
\begin{equation}
    v_{\text{OT}}(\mathbf{z}_t, t) = \int_{\mathbb{R}^m} (\mathbf{z}_1 - \mathbf{z}_0) \, d\pi^*(\mathbf{z}_0, \mathbf{z}_1 | \mathbf{z}_t).
\end{equation}
For discrete distributions, this formula becomes
\begin{equation}
    v_{\text{OT}}(\mathbf{z}_t, t) = \sum_{i,j} \pi^*_{ij} (\mathbf{z}_j^1 - \mathbf{z}_i^0) \delta(\mathbf{z}_t - \mathbf{z}_{t,ij}),
\end{equation}
where $\mathbf{z}_{t,ij} = (1-t)\mathbf{z}_i^0 + t\mathbf{z}_j^1$.

Optimizing three coupled manifolds simultaneously is unstable due to varying gradient scales. For example, small torsion changes can cause large Cartesian displacements. To ensure robust convergence, we introduce a three-stage curriculum training strategy, as shown in Appendix Algorithm 1.

At the first stage, we first train independent flow matching objectives for each subspace to learn their specific optimal transport structures:
\begin{align}
    \mathcal{L}_{sep} = \mathcal{L}_{trans}(\theta_T) + \mathcal{L}_{rot}(\theta_R) + \mathcal{L}_{conf}(\theta_C).
\end{align}
Here, $\mathcal{L}_{rot}$ minimizes the geodesic distance on $SO(3)$ , and $\mathcal{L}_{conf}$ minimizes the Wasserstein distance for internal coordinates.  We then jointly train the parameters $\theta$ with a flow consistency loss $\mathcal{L}_{flow}$ to ensure global coherence at the second stage:
\begin{align}
    \mathcal{L}_{total} = \sum_{k \in \{T,R,C\}} \lambda_k \mathcal{L}_k + \lambda_F \mathcal{L}_{flow},
\end{align}
\begin{align}
    \mathcal{L}_{flow} = \mathbb{E}[||v_{total}(X_t, t) - \dot{X}^{target}_t||^2],
\end{align}
where $\dot{X}^{target}_t$ is obtained by differentialing the composed manifold path $X_t = f(c_t, q_t,z_t): \dot{X}^{target} = \frac{\partial f}{\partial c}\dot{c_t} + \frac{\partial f}{\partial q}\dot{q_t} + \frac{\partial f}{\partial z}\dot{z_t}$.
The adaptive weighting $\lambda_k = 1/2\sigma_k^2$ based on homoscedastic uncertainty is employed to balance the loss components.
Finally, if likelihood-based fine-tuning is used, we treat the learned vector field as a continuous normalizing flow and optimize the data likelihood through the instantaneous change-of-variables formula.
\begin{equation}
    \frac{d\mathbf{X}}{dt} = v_\theta(\mathbf{X}_t, t), \frac{d \log{p_t(X_t)}}{dt} = - \text{Tr} (\frac{\partial{v_\theta}(X_t,t)}{\partial X_t}).
\end{equation}
The trace term can be estimated using the Hutchinson estimator: $\text{Tr} (\frac{\partial {v_\theta}}{\partial X_t}) \approx \mathbb{E}_{\epsilon \sim \mathcal{N}(0,I)}[\epsilon^{\top}\frac{\partial {v_\theta}}{\partial X_t}\epsilon]$. This term is used only for log-density evaluation, not for modifying the sample trajectory.

\subsection{Inference}
During inference, we sample $x_0 \sim \mathcal{N}(0, I)$ and integrate the learned combined velocity field using an adaptive ODE solver. 
\begin{equation}
    \frac{d\mathbf{x}}{dt} = v_\theta(\mathbf{x}_t, t), \quad \mathbf{x}_0 \sim \mathcal{N}(0, I).
\end{equation}
By leveraging the linearity of the optimal transport paths in our decomposed spaces, GO-Flow achieves high-fidelity generation with as few as 50 solver steps, significantly outperforming traditional diffusion methods that require thousands of steps.

%% file: sections/experiments.tex
\section{Experiments}

In this section, we present a series of comprehensive experiments designed to illustrate the efficacy of GO-Flow. These experiments include performance comparisons across standard molecular conformation generation benchmarks and an analysis of computational efficiency. To assess the impact of our manifold-aware framework, we conducted comparative analyses against state-of-the-art baselines, including both diffusion-based models and flow matching models such as CVGAE~\cite{mansimov2019molecular}, GraphDG~\cite{simm2019generative}, CGCF~\cite{xu2021learning}, ConfVAE~\cite{xu2021end}, 
GeoMol~\citep{ganea2021geomol}, 
ConfGF~\citep{shi2021learning}, GeoDiff~\citep{xu2022geodiff}, EBD~\cite{park2024equivariant}, MCF~\cite{wang2024swallowing}, ET-Flow~\cite{hassan2024flow} and AvgFlow~\cite{cao2025efficient}.

\subsection{Experimental Setup}
Following previous generative modeling benchmarks, GO-Flow is evaluated on the GEOM-Drugs and GEOM-QM9 datasets~\citep{axelrod2022geom}. GEOM-Drugs contains larger, drug-like molecules with complex geometries, while GEOM-QM9 consists of smaller molecules with quantum chemical properties. The datasets are processed to extract ground truth centre-of-mass, orientation quaternions, and internal coordinates (bond lengths, angles, and torsions) to supervise the manifold-aware training objectives.
We provide implementation details in Appendix \ref{app:details}.

\begin{table*}[htp]
  \caption{Geometric evaluation on GEOM-QM9 ($\delta = 0.5\text{\r{A}}$).}
  \label{tab:rmsd_qm9}
  \centering
  \vspace{-5pt}
  \resizebox{0.8\textwidth}{!}{
    \begin{tabular}{l|c|cccc|cccc}
    \toprule[1.0pt]
    & & \multicolumn{2}{c}{\shortstack[c]{COV-R ($\%$) $\uparrow$}}  & \multicolumn{2}{c|}{\shortstack[c]{MAT-R($\text{\r{A}}$) $\downarrow$}}  & \multicolumn{2}{c}{\shortstack[c]{COV-P ($\%$) $\uparrow$}}  & \multicolumn{2}{c}{\shortstack[c]{MAT-P ($\text{\r{A}}$) $\downarrow$}} \\
    Models & steps & Mean & Med & Mean & Med & Mean & Med & Mean & Med \\
    \midrule[0.8pt]
    RDKit& -  & 88.34 & \textbf{95.08} & 0.3544 & 0.2974 & \textbf{83.42} & \textbf{88.17} & 0.3747 & 0.3692 \\ 
    CVGAE& -  & 0.09 & 0.00 & 1.6713 & 1.6088 & - & - & - & - \\ 
    GraphDG& -  & 73.33 & 84.21 & 0.4245 & 0.3973 & 43.90 & 35.33 & 0.5809 & 0.5823 \\ 
    CGCF& -  & 78.05 & 82.48 & 0.4219 & 0.3900 & 36.49 & 33.57 & 0.6615 & 0.6427 \\
    ConfVAE& -  & 77.84 & 88.20 & 0.4154 & 0.3739 & 38.02 & 34.67 & 0.6215 & 0.6091 \\
    GeoMol& -  & 71.26 & 72.00 & 0.3731 & 0.3731 & - & - & - & - \\ 
    ConfGF& -  & 88.49 & 94.31 & 0.2673 & 0.2685 & 46.43 & 43.41 & 0.5224 & 0.5124 \\
    \midrule
    GeoDiff& 5000 & 88.02 & 92.33 & 0.2199 & 0.2116 & 53.72 & 52.36 & 0.4362 & 0.4259 \\
    MCF & 50 & 66.18 & 70.85 & 0.7826 & 0.6493 & 40.95 & 46.19 & 0.5730 & 0.5514 \\
    AvgFlow & 50 & 88.77 & 91.42 & 0.2217 & 0.2049 & 68.56 & 69.47 & 0.3775 & 0.3688 \\
    EBD & 50 & 89.37 & 93.21 & 0.2374 & 0.1903 & 61.31 & 60.46 & 0.3622 & 0.3517 \\ 
    ET-Flow & 50 & 89.51 & 94.26 & 0.2130 & 0.1892 & 67.36 & 69.54 & 0.3701 & 0.3649 \\
    \textbf{GO-Flow} & 50  & \textbf{90.08} & 94.57 & \textbf{0.2086} & \textbf{0.1845} & 69.42 & 70.91 & \textbf{0.3587} & \textbf{0.3462} \\ 
    \bottomrule[1.0pt]
    \end{tabular}
    }
\end{table*}

\begin{figure*}[htp]
    \centering
    \includegraphics[width=1\linewidth]{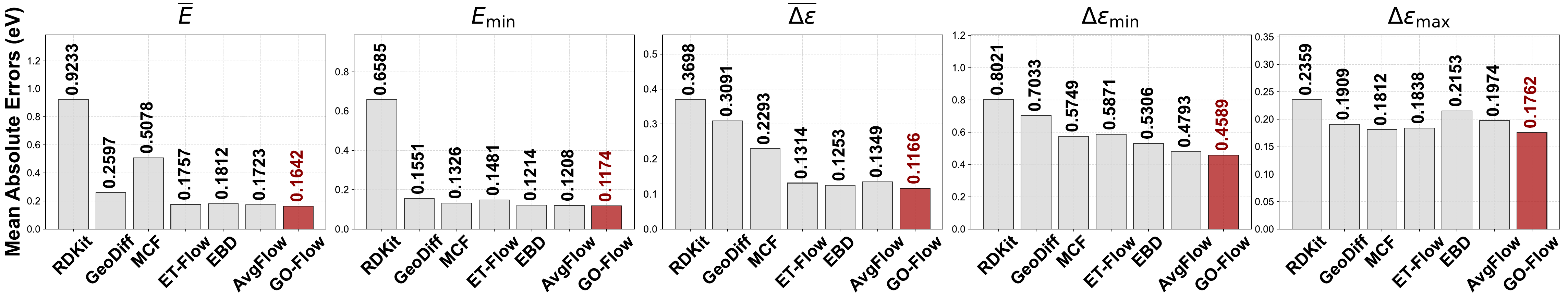}
    \caption{Mean absolute errors (MAE) between generated and ground truth ensemble properties in eV.}
    \vspace{-5pt}
    \label{fig:property}
\end{figure*}

\subsection{Performance on Molecular Conformation Generation}

We evaluated the generative performance of GO-Flow on two standard benchmark datasets: GEOM-Drugs and GEOM-QM9~\citep{axelrod2022geom}. Following established protocols~\cite{shi2021learning,xu2022geodiff,park2024equivariant}, we assessed the quality of the generated ensembles using Coverage (COV) and Matching (MAT) scores for both Recall (R) and Precision (P). These metrics collectively measure the diversity of the generated conformers and their geometric fidelity to the ground truth distributions.

\subsubsection{\textbf{Evaluation on GEOM-Drugs}}
The GEOM-Drugs dataset, characterized by large, drug-like molecules with complex topologies, serves as a rigorous testbed for modeling high-dimensional conformational spaces. Based on the experimental results shown in Table \ref{tab:rmsd_drug}, GO-Flow achieves state-of-the-art performance across all evaluation metrics, demonstrating a superior balance between diversity (Recall) and geometric quality (Precision). Specifically, GO-Flow attains a mean COV-R of 94.82\% and a median COV-R of 99.26\%. These results surpass the strong diffusion baseline GeoDiff (mean COV-R 89.40\%) and the flow-matching baseline ET-Flow (mean COV-R 93.12\%). The high coverage scores indicate that our manifold-aware framework effectively captures the multimodal nature of the probability distribution, successfully generating diverse conformational modes that cover the ground truth distribution more comprehensively than existing methods. A common challenge for generative models is maintaining structural validity while ensuring diversity. Standard Cartesian-based methods often struggle with invalid local structures due to unconstrained noise. In contrast, GO-Flow exhibits exceptional fidelity, recording the highest mean COV-P of 70.13\% and median COV-P of 72.49\%. This represents a significant margin over GeoDiff (mean COV-P 61.28\%) and AvgFlow (mean COV-P 68.56\%). Furthermore, our method achieves the lowest mean MAT-R score of 0.7971 \AA, indicating that the generated conformers are geometrically closer to the reference structures than those produced by any other baseline. This high precision validates the effectiveness of our manifold decomposition strategy, which strictly constrains generation to physically valid subspaces (e.g., $SO(3)$ for rotation and optimal transport for conformation), thereby minimizing the production of physically implausible outliers. Remarkably, GO-Flow achieves these superior results with significantly improved sampling efficiency. While diffusion-based models like GeoDiff require 5,000 steps to generate high-quality samples, GO-Flow converges to better performance with only 50 ODE solver steps.

\begin{table*}[htp]
  \caption{Ablation results on GEOM-Drugs and GEOM-QM9 with mean value of metrics.}
  \label{tab:abl}
  \vspace{-5pt}
  \centering
  \resizebox{0.95\textwidth}{!}{
    \begin{tabular}{l|cccc|cccc}
    \toprule[1.0pt]
    & \multicolumn{4}{c|}{\shortstack[c]{GEOM-Drugs}}  & \multicolumn{4}{c}{\shortstack[c]{GEOM-QM9}}   \\
    \#Components & COV-R ($\%$) $\uparrow$ & MAT-R($\text{\r{A}}$) $\downarrow$  & COV-P ($\%$) $\uparrow$ & MAT-P($\text{\r{A}}$) $\downarrow$ & COV-R ($\%$) $\uparrow$ & MAT-R($\text{\r{A}}$) $\downarrow$  & COV-P ($\%$) $\uparrow$ & MAT-P($\text{\r{A}}$) $\downarrow$ \\
    \midrule[0.8pt]    
    \textit{w/o} T &	94.05&	0.8146	&69.91&	1.1092&	90.46&	0.2095&	69.11&	0.3688   \\
    \textit{w/o} R &	91.41	&0.8367&	65.57&	1.1401&	89.14&	0.2159&	61.04&	0.4001   \\
    \textit{w/o} C &	90.26	&0.8512	&62.47	&1.1531	&88.06	&0.2149	&53.86	&0.4367  \\
    \midrule
    GO-Flow  & \textbf{94.82} & \bf 0.7971 & \textbf{70.13} & \textbf{1.1068}  & \textbf{90.08} & \bf 0.2086 & \bf 69.42 & \textbf{0.3587}\\ 
    \bottomrule[1.0pt]
    \end{tabular}
    }
\end{table*}


\begin{figure}[htp]
    \centering
    \includegraphics[width=1\linewidth]{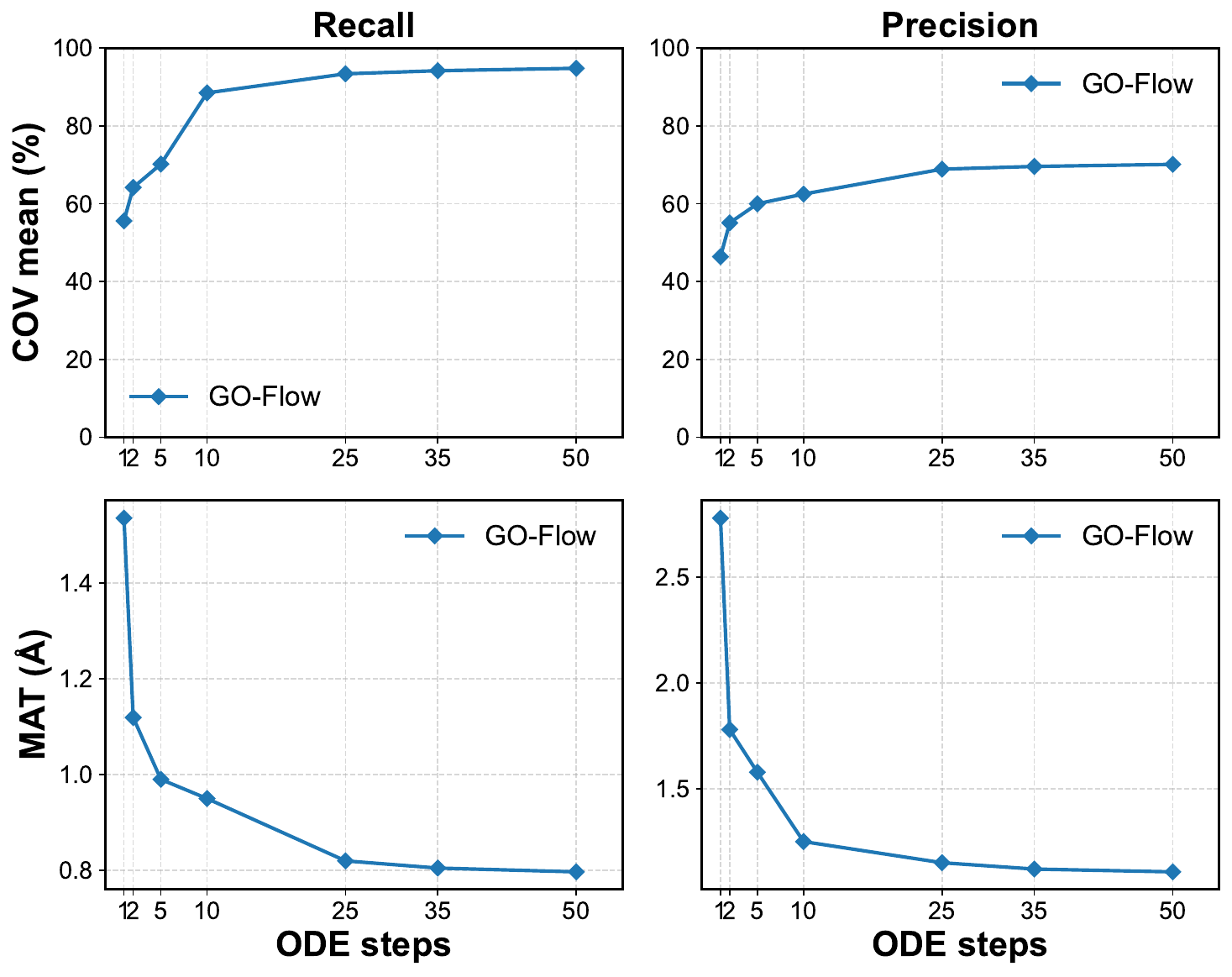}
    \vspace{-20pt}
    \caption{Effect of the number of ODE steps on model’s performance on GEOM-Drugs. }
    \vspace{-10pt}
    \label{fig:ODEnumber}
\end{figure}

\subsubsection{\textbf{Evaluation on GEOM-QM9}}
We further assessed the model on the GEOM-QM9 dataset to evaluate its precision in modeling smaller molecules where subtle geometric variations significantly impact quantum chemical properties. As shown in Table \ref{tab:rmsd_qm9}, GO-Flow demonstrates exceptional geometric accuracy. It achieves the lowest mean MAT-R score of 0.2086 \AA ~among all compared methods, improving upon the 0.2116 \AA ~of GeoDiff and substantially outperforming classical methods like RDKit (0.2974 \AA). A lower matching score implies that for any given ground truth conformer, GO-Flow can generate a sample that is geometrically nearly identical, which is crucial for tasks requiring high structural precision. Despite the high precision requirements, GO-Flow maintains competitive diversity. It achieves a mean COV-R of 90.08\% and a median COV-R of 94.57\%, outperforming AvgFlow (mean COV-R 88.77\%). Notably, this comprehensive coverage is achieved with only 50 sampling steps, whereas GeoDiff requires 5,000 steps to reach a lower mean COV-R of 88.02\%. This demonstrates that GO-Flow is not only more accurate but also significantly more efficient in exploring the conformational space of quantum chemical systems.

\subsection{Performance on Molecular Property Prediction}
\begin{figure*}[htp]
    \centering
    \vspace{-2cm}
    \includegraphics[width=1\linewidth]{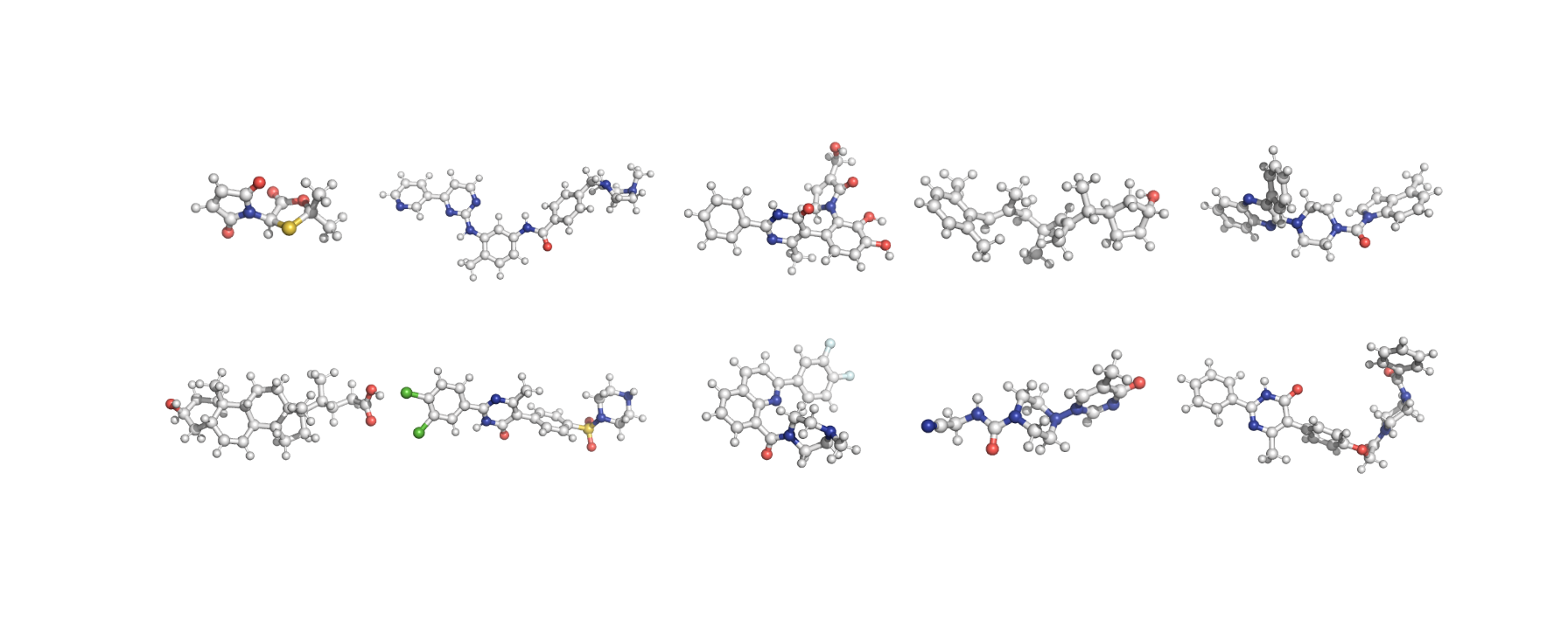}
    \vspace{-2.2cm}
    \caption{Representative samples generated by our model on GEOM-Drugs. It shows that GO-Flow produces globally coherent structures that maintain rigid scaffolds while generating diverse and physically plausible side-chain arrangements. }
    \label{fig:geom-Drugs}
\end{figure*}

Beyond geometric metrics, we further assessed the physical validity of the generated conformers by evaluating their thermodynamic and electronic properties.
We use PSI4~\cite{smith2020psi4} to calculate
properties of each conformer.
Figure ~\ref{fig:property} reports the mean absolute errors (MAE) between the properties of the generated ensembles and the ground truth, measured in electron-volts (eV). The evaluated properties include the total energy ($E$) and the HOMO-LUMO gap ($\Delta \epsilon$), along with their minimum and maximum values within the generated ensembles ($E_{\min}$, $\Delta \epsilon_{\min}$, $\Delta \epsilon_{\max}$). As shown in Figure ~\ref{fig:property}, GO-Flow achieves the lowest MAE across all five reported metrics, significantly outperforming both classical methods (e.g., RDKit) and advanced generative baselines including diffusion (GeoDiff) and other flow matching models (e.g., ET-Flow, AvgFlow). Specifically, for the mean total energy prediction ($\bar{E}$), GO-Flow attains an MAE of 0.1642 eV, which corresponds to a 36.8\% error reduction compared to GeoDiff (0.2597 eV) and also surpasses the best flow matching baseline (AvgFlow, 0.1723 eV). Furthermore, GO-Flow demonstrates superior precision in predicting electronic properties, achieving an MAE of 0.1166 eV for the HOMO-LUMO gap ($\Delta \epsilon$), strictly outperforming EBD (0.1253 eV) and ET-Flow (0.1314 eV). Since molecular potential energy is highly sensitive to subtle variations in bond lengths and angles, these results strongly evidence that our manifold decomposition strategy, particularly the precise modeling of internal coordinates via entropic optimal transport, yields chemically plausible structures that reside in low-energy regions of the potential energy surface. The reduced errors in extremal values, such as the $E_{\min}$ MAE of 0.1174 eV and $\Delta \epsilon_{\min}$ MAE of 0.4589 eV, indicate that GO-Flow generates diverse ensembles that are consistently valid, rather than merely producing a few high-quality samples amidst noisy ones.

\subsection{Ablation Study}

To verify the efficacy of each component in our manifold decomposition framework, we conducted a comprehensive ablation study on the GEOM-Drugs and GEOM-QM9 datasets. We evaluated three variants of GO-Flow by selectively removing the flow matching objectives associated with specific manifolds: translation ($\mathcal{M}_T$), rotation ($\mathcal{M}_R$), and conformation ($\mathcal{M}_C$). We report the mean value of metrics. The results are summarized in Table ~\ref{tab:abl}.

\subsubsection{\textbf{Impact of Conformation Manifold (\textit{w/o} C)}}
The removal of the conformation manifold decomposition yields the most significant performance degradation, particularly on the complex GEOM-Drugs dataset. As shown in Table 3, the w/o C variant achieves a COV-R of only 90.26\% and a MAT-R of 0.8512 \AA, which is substantially inferior to the full GO-Flow model (94.82\% and 0.7971 \AA, respectively). This variant essentially degrades to a Cartesian-based diffusion model similar to GeoDiff~\cite{xu2022geodiff}, failing to explicitly capture the internal geometric constraints. These results underscore that modeling internal coordinates (bond lengths, angles, and torsions) via entropic optimal transport is the most critical factor for generating high-fidelity molecular structures.

\subsubsection{\textbf{Impact of Rotation Manifold (\textit{w/o} R)}}
Excluding the rotation manifold also leads to a noticeable drop in generation quality. The \textit{w/o} R variant, which treats orientation in Euclidean space rather than on the $SO(3)$ manifold, results in a COV-R decrease to 91.41\% on GEOM-Drugs. While it outperforms the conformation-agnostic baseline (\textit{w/o} C), it still lags behind the full model. This suggests that respecting the Lie group structure of rotations is essential for maintaining global structural coherence and equivariance, which linear approximations cannot fully capture.

\subsubsection{\textbf{Impact of Translation Manifold (\textit{w/o} T)}}
The\textit{ w/o} T variant, which removes the specific modeling of the center-of-mass translation, exhibits the smallest performance gap compared to the full model (e.g., 94.05\% COV-R on GEOM-Drugs). This outcome is expected, as translation represents the simplest degree of freedom in molecular generation. However, the full GO-Flow model still consistently achieves the best metrics across all tasks, indicating that a unified framework treating all geometric aspects, including translation, as flow matching problems on their respective manifolds contributes to optimal stability and convergence.

\subsubsection{\textbf{Effect of the number of ODE steps}}

To investigate the convergence behavior and inference efficiency of GO-Flow, we evaluated the generation quality on the GEOM-Drugs dataset across varying numbers of ODE solver steps $T \in \{1, 2, 5, 10, 25, 35, 50\}$. As illustrated in Figure ~\ref{fig:ODEnumber}, GO-Flow exhibits a rapid convergence rate, demonstrating the effectiveness of our manifold-aware trajectory design. 

Even with an extremely limited computational budget, GO-Flow produces chemically plausible structures. Notably, at just $T=10$ steps, the model achieves a COV-R of 88.5\%, which is already comparable to the performance of the diffusion-based baseline GeoDiff at 5000 steps ($89.40\%$). As the number of steps increases to 25, the performance further improves to a COV-R of 93.4\% and a MAT-R of 0.82 $\text{\AA}$, effectively surpassing the performance of recent flow matching baselines like AvgFlow ($91.72\%$ COV-R at $T=50$). The performance metrics largely plateau after 35 steps, indicating that the model successfully identifies the high-density regions of the Boltzmann distribution without requiring excessive integration steps. While standard diffusion models often require thousands of steps to resolve geometric dependencies, GO-Flow reaches state-of-the-art performance (COV-R 94.82\%, MAT-R 0.7971 $\text{\AA}$) at $T=50$.

\subsection{Qualitative Visualization of Generated Conformers}

We provide a visual inspection of the generated samples to qualitatively assess their chemical validity and structural diversity.

Figure~\ref{fig:geom-Drugs} illustrates generated samples from the more challenging GEOM-Drugs dataset. These molecules are characterized by larger sizes and higher degrees of torsional flexibility. Despite this complexity, GO-Flow generates globally coherent structures. The figure demonstrates the model's capability to navigate the high-dimensional conformational space, producing diverse and physically plausible arrangements of flexible side chains and rigid scaffolds. This aligns with the high precision scores (COV-P) reported in Table~\ref{tab:rmsd_drug}, confirming that our manifold decomposition strategy effectively scales to large, drug-like systems without sacrificing geometric consistency. We also provide visualisation on small molecule (GEOM-QM9) in Appendix \ref{app:visualization_qm9}.



%% file: sections/conclusion.tex
\section{Conclusion}

In this paper, we presented GO-Flow, a manifold-aware flow matching framework that fundamentally rethinks 3D molecular generation. By departing from the prevailing Cartesian-centric paradigm and decomposing molecular motion into physically distinct manifolds, including translation, rotation, and conformation, GO-Flow effectively integrates geometric inductive biases directly into the generative process. We designed rigorous flow matching objectives, including geodesic flows on $SO(3)$ and Entropic Optimal Transport for internal coordinates, to strictly enforce equivariance and handle the complex energy landscape of molecular conformations.
Our empirical results on standard benchmarks demonstrate that GO-Flow not only matches or outperforms state-of-the-art baselines in generation quality but also exhibits superior geometric consistency. Crucially, the significant efficiency gain, requiring only 50 solver steps, is not merely an engineering optimization, but a theoretical consequence of modeling probability paths that respect the underlying geometry of the data. We believe GO-Flow offers a promising direction for AI-driven drug discovery, suggesting that respecting physical manifolds is a prerequisite for building scalable and interpretable generative models for scientific data.

\section{Limitations and Ethical Considerations}
As a conformation generation model, GO-Flow operates on a predefined 2D molecular graph. It generates 3D structures for a known chemical formula but does not currently support de novo design tasks, such as generating atoms and bonds from scratch or modifying the molecular topology. Integrating our manifold-aware flow matching with graph generative models to strictly couple chemical and conformational space exploration remains a promising direction for future research.

This work focuses on 3D molecular conformation generation, a fundamental task in drug discovery and materials science. The primary societal impact is the potential to expedite the development of new therapeutics, benefiting public health. 
However, as with any generative technology in chemistry, there is a theoretical risk of misuse for designing harmful substances. We believe this risk is mitigated by the fact that our model requires a known molecular graph input and does not autonomously propose toxic compounds.
All data used in this work (GEOM-Drugs, GEOM-QM9) are publicly available and contain no personally identifiable information.

\section*{GenAI Disclosure}

The authors utilized a GenAI solely as a tool to assist with the polishing and refinement of the writing in this paper. The model was used exclusively for improving grammatical fluency, sentence structure, and overall clarity of the manuscript. All ideation, theoretical development, empirical research, and technical conclusions remain entirely the work of the authors. The authors take full responsibility for all content generated by the GenAI and presented in this work.

%% file: sections/appendix.tex
\clearpage
\appendix

\paragraph{Table of Appendix:}
\begin{itemize}
    \item \ref{app:A}. Coordinate Space Decomposition
    \begin{itemize}
        \item \ref{app:A.translation}. Translation Space
        \item \ref{app:A.rot}. Rotation Space
        \item \ref{app:A.conf}. Conformation Space
        \item \ref{app:velco}. Velocity Field Composition
        
    \end{itemize}
    \item \ref{app:loss}. Loss Function and Flow Matching Objective
    \begin{itemize}
        \item \ref{app:trans_loss}. Translation Flow Matching Loss
        \item \ref{app:rot_loss}. Rotation Flow Matching Loss
        \item \ref{app:conf_loss}. Conformation Flow Matching Loss
        \item \ref{app:comb_loss}. Combined Loss with Adaptive Weighting
        \item \ref{app:convergence}. Three-Stage Training Strategy with Convergence Analysis
    \end{itemize}
    \item \ref{app:derivations}.  Mathematical Derivations
    \begin{itemize}
        \item \ref{app:SLERP}. Quaternion SLERP Analysis
        \item \ref{app:Jacobian}. Internal Coordinate Jacobian
        \item \ref{app:OT}. Optimal Transport for Molucular Conformations
        \item \ref{app:FM}. Flow Matching
    \end{itemize}
    \item \ref{app:details}. Implementation Details
    \begin{itemize}
        \item \ref{apdx:subsec_data}. Datasets
        \item \ref{app:input}. Input Featurization
        \item \ref{app:metrics}. Evaluation Metrics
        \item \ref{app:hyper}. Hyperparameters
    \end{itemize}
    \item \ref{app:visualization_qm9}. Visualization on Small Molecules (GEOM-QM9)
\end{itemize}

\section{Coordinate Space Decomposition}
\label{app:A}

\subsection{Translation Space}
\label{app:A.translation}
We use linear optimal transport in $\mathbb{R}^3$.
For the center of mass translation, the optimal transport problem is 
\begin{equation}
    \min_{\gamma \in \Pi(\mu_0, \mu_1)} \int_{\mathbb{R}^3 \times \mathbb{R}^3} \|\mathbf{c}_0 - \mathbf{c}_1\|_2^2 \, d\gamma(\mathbf{c}_0, \mathbf{c}_1),
\end{equation}
where $\Pi(\mu_0, \mu_1)$ is the set of couplings between source distribution $\mu_0$ and target distribution $\mu_1$.
For Gaussian distributions, the optimal transport map is linear 
\begin{equation}
    T^*(\mathbf{c}_0) = \mathbf{c}_0 + \boldsymbol{\mu}_1 - \boldsymbol{\mu}_0.
\end{equation}
This leads to the interpolation path 
\begin{equation}
    \mathbf{c}_t = (1-t)\mathbf{c}_0 + t\mathbf{c}_1, \quad t \in [0,1],
\end{equation}
where $\mathbf{c}_0 \sim \mathcal{N}(\boldsymbol{\mu}_0, \sigma^2I)$ and $\mathbf{c}_1$ is the target center of mass.
The velocity field is constant 
\begin{equation}
    v_{\text{trans}}(\mathbf{c}_t, t) = \mathbf{c}_1 - \mathbf{c}_0 = \boldsymbol{\mu}_1 - \boldsymbol{\mu}_0.
\end{equation}

\subsection{Rotation Space}
\label{app:A.rot}
We use geodesic flow on SO(3) manifold.
Molecular orientations form the Special Orthogonal group SO(3). We parameterize rotations using unit quaternions $\mathbf{q} \in \mathbb{S}^3 \subset \mathbb{R}^4$ to avoid singularities.
The geodesic distance on SO(3) between two rotations $\mathbf{R}_0, \mathbf{R}_1$ is 
\begin{equation}
    d_{\text{geo}}(\mathbf{R}_0, \mathbf{R}_1) = \|\log(\mathbf{R}_1 \mathbf{R}_0^T)\|_F,
\end{equation}
where $\log$ is the matrix logarithm and $\|\cdot\|_F$ is the Frobenius norm.
For quaternions, the geodesic path is computed via Spherical Linear Interpolation (SLERP)
\begin{equation}
    \mathbf{q}_t = \text{SLERP}(\mathbf{q}_0, \mathbf{q}_1, t) = \frac{\sin((1-t)\theta)}{\sin\theta}\mathbf{q}_0 + \frac{\sin(t\theta)}{\sin\theta}\mathbf{q}_1,
\end{equation}
where $\cos\theta = |\mathbf{q}_0 \cdot \mathbf{q}_1|$ (we take absolute value to ensure shortest path).

\subsubsection{Derivation of Angular Velocity} 
The angular velocity in the $\mathfrak{so}(3)$ Lie algebra is obtained from the relative quaternion 
\begin{align}
    \mathbf{q}_{\text{rel}} &= \mathbf{q}_1 \mathbf{q}_0^{-1} = \mathbf{q}_1 \mathbf{q}_0^* ,\\
    \boldsymbol{\omega} &= 2 \cdot \text{Im}(\log(\mathbf{q}_{\text{rel}})) \\
    &= 2 \cdot \frac{\text{Im}(\mathbf{q}_{\text{rel}})}{\|\text{Im}(\mathbf{q}_{\text{rel}})\|} \arccos(\text{Re}(\mathbf{q}_{\text{rel}})), 
\end{align}
where $\text{Im}(\mathbf{q})$ and $\text{Re}(\mathbf{q})$ are the imaginary and real parts of quaternion $\mathbf{q}$.
The velocity field for rotation is
\begin{equation}
    v_{\text{rot}}(\mathbf{q}_t, t) = \frac{d\mathbf{q}_t}{dt} = \frac{1}{2}\boldsymbol{\omega} \mathbf{q}_t.
\end{equation}

\subsection{Conformation Space}
\label{app:A.conf}
We use entropic optimal transport.
Internal coordinates $\mathbf{z} = \{r_{ij}, \theta_{ijk}, \phi_{ijkl}\}$ represent bond lengths, angles, and dihedrals. The conformational space is constrained by chemical validity.
We formulate the entropic optimal transport problem
\begin{equation}
    \min_{\pi \in \Pi(\mu_0, \mu_1)} \langle C(\mathbf{z}_0, \mathbf{z}_1), \pi \rangle + \epsilon H(\pi),
\end{equation}
where $C(z_0,z_1)
=
\alpha_r\|r_0-r_1\|_2^2
+
\alpha_\theta\|\theta_0-\theta_1\|_2^2
+
\alpha_\phi\|\operatorname{wrap}(\phi_0-\phi_1)\|_2^2,$ is the cost matrix, $\operatorname{wrap}(\Delta\phi)
=
\operatorname{atan2}(\sin\Delta\phi,\cos\Delta\phi)$, $H(\pi) = -\sum_{i,j} \pi_{ij} \log \pi_{ij}$ is the entropy regularization and $\epsilon > 0$ controls the trade-off between cost and entropy.

\subsubsection{Sinkhorn Algorithm}
The entropic optimal transport is solved iteratively
\begin{align}
    \mathbf{u}^{(k+1)} &= \frac{\mathbf{a}}{\mathbf{K} \mathbf{v}^{(k)}} ,\\
    \mathbf{v}^{(k+1)} &= \frac{\mathbf{b}}{\mathbf{K}^T \mathbf{u}^{(k+1)}}, 
\end{align}
where $\mathbf{K}_{ij} = \exp(-C_{ij}/\epsilon)$, $\mathbf{a}, \mathbf{b}$ are marginal distributions, and $\mathbf{u}, \mathbf{v}$ are dual variables.
The optimal transport plan is
\begin{equation}
    \pi^* = \text{diag}(\mathbf{u}) \mathbf{K} \text{diag}(\mathbf{v}).
\end{equation}
The velocity field is derived from the Wasserstein gradient flow
\begin{equation}
    v_{\text{conf}}(\mathbf{z}_t, t) = -\nabla_{\mathbf{z}} W_2^2(\mu_t, \mu_1),
\end{equation}
where $W_2$ is the 2-Wasserstein distance.

\subsection{Velocity Field Composition}
\label{app:velco}

For translation space, translation velocity contribution is
\begin{equation}
    v_{\text{trans}}^{\text{cart}} = v_{\text{trans}} \mathbf{1}_N, 
\end{equation}
where $\mathbf{1}_N$ replicates the center-of-mass velocity to all atoms.
For each molecule $m$ with atoms $\mathcal{A}_m$, the rotational velocity contribution is
\begin{equation}
    v_{\text{rot}}^{\text{cart}}(\mathbf{x}_i) = \boldsymbol{\omega}_m \times (\mathbf{x}_i - \mathbf{c}_m), \quad i \in \mathcal{A}_m, 
\end{equation}
where $\mathbf{c}_m$ is the center of mass of molecule $m$.
As for conformation space, the transformation from internal coordinate velocity to Cartesian requires the Jacobian 
\begin{equation}
    v_{\text{conf}}^{\text{cart}} = J(\mathbf{z}) v_{\text{conf}}, 
\end{equation}
where $J(\mathbf{z}) = \frac{\partial \mathbf{X}}{\partial \mathbf{z}} \in \mathbb{R}^{3N \times m}$ is the Jacobian matrix. For a bond length $r_{ij}$ 
\begin{equation}
    \frac{\partial \mathbf{x}_i}{\partial r_{ij}} = \frac{\mathbf{x}_i - \mathbf{x}_j}{\|\mathbf{x}_i - \mathbf{x}_j\|}, \quad \frac{\partial \mathbf{x}_j}{\partial r_{ij}} = -\frac{\mathbf{x}_i - \mathbf{x}_j}{\|\mathbf{x}_i - \mathbf{x}_j\|}.
\end{equation}
For a bond angle $\theta_{ijk}$ (atom $j$ is the center) 
\begin{align}
    \frac{\partial \mathbf{x}_i}{\partial \theta_{ijk}} &= \frac{(\mathbf{u}_{ji} \times \mathbf{u}_{jk}) \times \mathbf{u}_{ji}}{\|\mathbf{u}_{ji}\| \sin\theta_{ijk}} ,\\
    \frac{\partial \mathbf{x}_k}{\partial \theta_{ijk}} &= \frac{(\mathbf{u}_{jk} \times \mathbf{u}_{ji}) \times \mathbf{u}_{jk}}{\|\mathbf{u}_{jk}\| \sin\theta_{ijk}},
\end{align}
where $\mathbf{u}_{ab} = \mathbf{x}_a - \mathbf{x}_b$.
The total Cartesian velocity field is 
\begin{equation}
    v_{\text{total}}(\mathbf{X}, t) = v_{\text{trans}}^{\text{cart}} + v_{\text{rot}}^{\text{cart}} + v_{\text{conf}}^{\text{cart}}.
\end{equation}
For each coordinate space, we learn a velocity field $v_\theta  \mathbb{R}^d \times [0,1] \rightarrow \mathbb{R}^d$ 
\begin{align}
    &v_{\text{trans}}(\mathbf{c}_t, t)   \mathbb{R}^3 \times [0,1] \rightarrow \mathbb{R}^3 ,\\
    &v_{\text{rot}}(\mathbf{q}_t, t)   \mathbb{S}^3 \times [0,1] \rightarrow SO(3) ,\\
    &v_{\text{conf}}(\mathbf{z}_t, t)   \mathbb{R}^m \times [0,1] \rightarrow \mathbb{R}^m .
\end{align}
The combined velocity in Cartesian space is 
\begin{equation}
    v_{\text{total}} = v_{\text{trans}} + \boldsymbol{\omega} \times \mathbf{r} + J(\mathbf{z}) v_{\text{conf}},
\end{equation}
where $\mathbf{r}$ is the centered atomic position and $J(\mathbf{z})$ is the Jacobian from internal to Cartesian coordinates.

\section{Loss Function and Flow Matching Objective}
\label{app:loss}

The training objective combines flow matching losses from all three coordinate spaces. We derive each loss component from the corresponding optimal transport formulation.

\subsection{Translation Flow Matching Loss}
\label{app:trans_loss}
For linear optimal transport in translation space, the flow matching loss is
\begin{equation}
    \mathcal{L}_{\text{trans}} = \mathbb{E}_{t,\mathbf{c}_0,\mathbf{c}_1} \left[ \|v_{\text{trans}}(\mathbf{c}_t, t) - (\mathbf{c}_1 - \mathbf{c}_0)\|^2 \right] ,
\end{equation}
where $\mathbf{c}_t = (1-t)\mathbf{c}_0 + t\mathbf{c}_1$ and the target velocity is the constant vector $\mathbf{c}_1 - \mathbf{c}_0$.

\subsection{Rotation Flow Matching Loss}
\label{app:rot_loss}
For geodesic flow on SO(3), we use the geodesic distance between predicted and target angular velocities 
\begin{equation}
    \mathcal{L}_{\text{rot}} = \mathbb{E}_{t,\mathbf{q}_0,\mathbf{q}_1} \left[ d_{\text{geo}}^2(v_{\text{rot}}(\mathbf{q}_t, t), \boldsymbol{\omega}_{\text{target}}) \right],
\end{equation}
where
\begin{align}
    &\mathbf{q}_t = \text{SLERP}(\mathbf{q}_0, \mathbf{q}_1, t) ,\\
    &\boldsymbol{\omega}_{\text{target}} = 2 \cdot \text{Im}(\log(\mathbf{q}_1 \mathbf{q}_0^*)) .
\end{align}
The geodesic distance for velocities in $SO(3)$ is 
\begin{equation}
    d_{\text{geo}}(\boldsymbol{\omega}_1, \boldsymbol{\omega}_2) = \|\boldsymbol{\omega}_1 - \boldsymbol{\omega}_2\|_2.
\end{equation}

\subsection{Conformation Flow Matching Loss}
\label{app:conf_loss}
For optimal transport in internal coordinate space, we use the Wasserstein distance
\begin{equation}
    \mathcal{L}_{\text{conf}} = \mathbb{E}_{t,\mathbf{z}_0,\mathbf{z}_1} \left[ W_2^2(v_{\text{conf}}(\mathbf{z}_t, t), v_{\text{OT}}(\mathbf{z}_t, t)) \right].
\end{equation}
where $v_{\text{OT}}(\mathbf{z}_t, t)$ is the velocity field induced by the optimal transport plan $\pi^*$. 
The velocity field is computed as 
\begin{equation}
    v_{\text{OT}}(\mathbf{z}_t, t) = \int_{\mathbb{R}^m} (\mathbf{z}_1 - \mathbf{z}_0) \, d\pi^*(\mathbf{z}_0, \mathbf{z}_1 | \mathbf{z}_t).
\end{equation}
For discrete distributions, this formula becomes
\begin{equation}
    v_{\text{OT}}(\mathbf{z}_t, t) = \sum_{i,j} \pi^*_{ij} (\mathbf{z}_j^1 - \mathbf{z}_i^0) \delta(\mathbf{z}_t - \mathbf{z}_{t,ij}),
\end{equation}
where $\mathbf{z}_{t,ij} = (1-t)\mathbf{z}_i^0 + t\mathbf{z}_j^1$.

\subsection{Combined Loss with Adaptive Weighting}
\label{app:comb_loss}
The total loss combines all three components 
\begin{equation}
    \mathcal{L}_{\text{total}} = \lambda_T \mathcal{L}_{\text{trans}} + \lambda_R \mathcal{L}_{\text{rot}} + \lambda_C \mathcal{L}_{\text{conf}} + \lambda_F \mathcal{L}_{\text{flow}}.
\end{equation}
We include an additional flow consistency loss 
\begin{equation}
    \mathcal{L}_{\text{flow}} = \mathbb{E} \left[ \|v_{\text{total}}(\mathbf{X}_t, t) - v_{\text{target}}(\mathbf{X}_t, t)\|^2 \right],
\end{equation}
where $v_{\text{target}} = \mathbf{X}_1 - \mathbf{X}_0$ is the target Cartesian velocity.
We use uncertainty-based weighting 
\begin{equation}
    \lambda_k = \frac{1}{2\sigma_k^2}, \quad \mathcal{L}_{\text{uncertainty}} = \sum_k \log \sigma_k^2,
\end{equation}
where $\sigma_k^2$ are learnable parameters representing task-specific uncertainty.

\subsection{Three-Stage Training Strategy with Convergence Analysis}
\label{app:convergence}

Our training strategy progressively builds complexity while ensuring stable optimization, as shown in Algorithm 1.

\subsubsection{Stage 1}
Each coordinate space is trained independently to learn its optimal transport structure.
For translation training,
\begin{align}
    \theta_T^{(k+1)} &= \theta_T^{(k)} - \alpha_T \nabla_{\theta_T} \mathcal{L}_{\text{trans}}(\theta_T^{(k)}) ,\\
    \mathcal{L}_{\text{trans}}(\theta_T) &= \mathbb{E}_{t,\mathbf{c}_0,\mathbf{c}_1} \left[ \|v_{\text{trans}}(\mathbf{c}_t, t; \theta_T) - (\mathbf{c}_1 - \mathbf{c}_0)\|^2 \right].
\end{align}
For rotation training,
\begin{align}
    \theta_R^{(k+1)} &= \theta_R^{(k)} - \alpha_R \nabla_{\theta_R} \mathcal{L}_{\text{rot}}(\theta_R^{(k)}) ,\\
    \mathcal{L}_{\text{rot}}(\theta_R) &= \mathbb{E}_{t,\mathbf{q}_0,\mathbf{q}_1} \left[ d_{\text{geo}}^2(v_{\text{rot}}(\mathbf{q}_t, t; \theta_R), \boldsymbol{\omega}_{\text{target}}) \right].
\end{align}
As for conformation training, 
\begin{align}
    \theta_C^{(k+1)} &= \theta_C^{(k)} - \alpha_C \nabla_{\theta_C} \mathcal{L}_{\text{conf}}(\theta_C^{(k)}) ,\\
    \mathcal{L}_{\text{conf}}(\theta_C) &= \mathbb{E}_{t,\mathbf{z}_0,\mathbf{z}_1} \left[ \|v_{\text{conf}}(\mathbf{z}_t, t; \theta_C) - v_{\text{OT}}(\mathbf{z}_t, t)\|^2 \right].
\end{align}
Each coordinate space has a unique global minimum due to the convex nature of optimal transport objectives.

\subsubsection{Stage 2}
Initialize with Stage 1 weights and train jointly 
\begin{align}
    \theta^{(k+1)} &= \theta^{(k)} - \alpha \nabla_\theta \mathcal{L}_{\text{total}}(\theta^{(k)}) ,\\
    \mathcal{L}_{\text{total}}(\theta) &= \lambda_T \mathcal{L}_{\text{trans}}(\theta_T) + \lambda_R \mathcal{L}_{\text{rot}}(\theta_R) + \lambda_C \mathcal{L}_{\text{conf}}(\theta_C) + \lambda_F \mathcal{L}_{\text{flow}}(\theta),
\end{align}
where $\theta = \{\theta_T, \theta_R, \theta_C\}$ are the joint parameters.
The flow consistency loss $\mathcal{L}_{\text{flow}}$ couples the coordinate spaces 
\begin{equation}
    \frac{\partial \mathcal{L}_{\text{flow}}}{\partial \theta_i} = \mathbb{E} \left[ 2(v_{\text{total}} - v_{\text{target}}) \cdot \frac{\partial v_{\text{total}}}{\partial \theta_i} \right].
\end{equation}
This creates gradient coupling between coordinate spaces, enabling coordinated learning.

\subsubsection{Stage 3}
Fine-tune with the probability flow ODE 
\begin{equation}
    \frac{d\mathbf{X}}{dt} = v_\theta(\mathbf{X}_t, t) - \frac{1}{2}\nabla \cdot v_\theta(\mathbf{X}_t, t).
\end{equation}

The divergence term $\nabla \cdot v_\theta$ is computed using the Hutchinson trace estimator 
\begin{equation}
    \nabla \cdot v_\theta(\mathbf{X}, t) \approx \mathbb{E}_{\boldsymbol{\epsilon} \sim \mathcal{N}(0,I)} \left[ \boldsymbol{\epsilon}^T \nabla v_\theta(\mathbf{X}, t) \boldsymbol{\epsilon} \right].
\end{equation}

\subsubsection{Sampling}

At inference time, we solve the ODE 
\begin{equation}
    \frac{d\mathbf{x}}{dt} = v_\theta(\mathbf{x}_t, t), \quad \mathbf{x}_0 \sim \mathcal{N}(0, I), 
\end{equation}
using adaptive ODE solvers (e.g., Dormand-Prince) with only 10-50 steps instead of 1000+ steps required by diffusion models.

\section{Mathematical Derivations}
\label{app:derivations}

\subsection{Quaternion SLERP Analysis}
\label{app:SLERP}
Given two unit quaternions $\mathbf{q}_0, \mathbf{q}_1 \in \mathbb{S}^3$, the shortest geodesic path is 
\begin{equation}
    \mathbf{q}(t) = \mathbf{q}_0 \left(\mathbf{q}_0^{-1} \mathbf{q}_1\right)^t,
\end{equation}
where $\mathbf{q}^t$ denotes quaternion exponentiation.
For the interpolation parameter $t \in [0,1]$, this becomes 
\begin{equation}
    \text{SLERP}(\mathbf{q}_0, \mathbf{q}_1, t) = \frac{\sin((1-t)\Omega)}{\sin\Omega}\mathbf{q}_0 + \frac{\sin(t\Omega)}{\sin\Omega}\mathbf{q}_1,
\end{equation}
where $\cos\Omega = \mathbf{q}_0 \cdot \mathbf{q}_1$.
The angular velocity is obtained by differentiating the SLERP 
\begin{align}
    \boldsymbol{\omega}(t) &= 2 \cdot \text{Im}\left(\frac{d\mathbf{q}}{dt} \mathbf{q}^{-1}\right) \\
    &= 2 \cdot \text{Im}\left(\frac{\Omega}{\sin\Omega}\left[-\cos((1-t)\Omega)\mathbf{q}_0 + \cos(t\Omega)\mathbf{q}_1\right] \mathbf{q}(t)^{-1}\right).
\end{align}
For constant angular velocity (which occurs in our optimal transport formulation) 
\begin{equation}
    \boldsymbol{\omega}_{\text{const}} = 2 \cdot \text{Im}\left(\log\left(\mathbf{q}_1 \mathbf{q}_0^{-1}\right)\right).
\end{equation}

\begin{table*}[htb]
  \centering
  \caption{Atomic features included in GO-flow.}
  \label{tab:atomic}
  \small
  \begin{tabular}{l l l}
    \toprule
    Name & Description & Range \\
    \midrule
    \texttt{chirality} & Chirality Tag & \{unspecified, tetrahedral CW \& CCW, other\} \\
    \texttt{degree} & Number of bonded neighbors & $\{x:0 \leq x \leq 10, x \in \mathbb{Z}\}$ \\
    \texttt{charge} & Formal charge of atom & $\{x:-5 \leq x \leq 5, x \in \mathbb{Z}\}$ \\
    \texttt{num\_H} & Total Number of Hydrogens & $\{x:0 \leq x \leq 8, x \in \mathbb{Z}\}$ \\
    \texttt{number\_radical\_e} & Number of Radical Electrons & $\{x:0 \leq x \leq 4, x \in \mathbb{Z}\}$ \\    
    \texttt{hybrization} & Hybrization type & \{sp, sp\textsuperscript{2}, sp\textsuperscript{3}, sp\textsuperscript{3}d, sp\textsuperscript{3}d\textsuperscript{2}, other\} \\
    \texttt{aromatic} & Whether on a aromatic ring & \{True, False\} \\
    \texttt{in\_ring} & Whether in a ring & \{True, False\} \\
    \bottomrule
  \end{tabular}
\end{table*}

\subsection{Internal Coordinate Jacobian}
\label{app:Jacobian}

\subsubsection{Bond Length Jacobian}
For a bond length $r_{ij} = \|\mathbf{x}_i - \mathbf{x}_j\|$ 
\begin{align}
    \frac{\partial r_{ij}}{\partial \mathbf{x}_i} &= \frac{\mathbf{x}_i - \mathbf{x}_j}{\|\mathbf{x}_i - \mathbf{x}_j\|} ,\\
    \frac{\partial r_{ij}}{\partial \mathbf{x}_j} &= -\frac{\mathbf{x}_i - \mathbf{x}_j}{\|\mathbf{x}_i - \mathbf{x}_j\|} ,\\
    \frac{\partial r_{ij}}{\partial \mathbf{x}_k} &= \mathbf{0}, \quad \text{for } k \neq i,j.
\end{align}

\subsubsection{Bond Angle Jacobian}
For a bond angle $\theta_{ijk}$ with vertex at atom $j$ 
\begin{align}
    \cos\theta_{ijk} &= \frac{(\mathbf{x}_i - \mathbf{x}_j) \cdot (\mathbf{x}_k - \mathbf{x}_j)}{\|\mathbf{x}_i - \mathbf{x}_j\| \|\mathbf{x}_k - \mathbf{x}_j\|}.
\end{align}
Define $\mathbf{u} = \frac{\mathbf{x}_i - \mathbf{x}_j}{\|\mathbf{x}_i - \mathbf{x}_j\|}$ and $\mathbf{v} = \frac{\mathbf{x}_k - \mathbf{x}_j}{\|\mathbf{x}_k - \mathbf{x}_j\|}$. Then 
\begin{align}
    \frac{\partial \theta_{ijk}}{\partial \mathbf{x}_i} &= \frac{1}{\sin\theta_{ijk}} \cdot \frac{1}{\|\mathbf{x}_i - \mathbf{x}_j\|} \left(\mathbf{v} - \mathbf{u}(\mathbf{u} \cdot \mathbf{v})\right) ,\\
    \frac{\partial \theta_{ijk}}{\partial \mathbf{x}_k} &= \frac{1}{\sin\theta_{ijk}} \cdot \frac{1}{\|\mathbf{x}_k - \mathbf{x}_j\|} \left(\mathbf{u} - \mathbf{v}(\mathbf{u} \cdot \mathbf{v})\right) ,\\
    \frac{\partial \theta_{ijk}}{\partial \mathbf{x}_j} &= -\frac{\partial \theta_{ijk}}{\partial \mathbf{x}_i} - \frac{\partial \theta_{ijk}}{\partial \mathbf{x}_k}.
\end{align}

\subsubsection{Dihedral Angle Jacobian}
For a dihedral angle $\phi_{ijkl}$ defined by four atoms 
\begin{align}
    \cos\phi_{ijkl} &= \frac{\mathbf{n}_1 \cdot \mathbf{n}_2}{\|\mathbf{n}_1\| \|\mathbf{n}_2\|} ,\\
    \sin\phi_{ijkl} &= \frac{(\mathbf{n}_1 \times \mathbf{n}_2) \cdot \mathbf{b}}{\|\mathbf{n}_1\| \|\mathbf{n}_2\| \|\mathbf{b}\|},
\end{align}
where
\begin{align}
    \mathbf{b} &= \mathbf{x}_k - \mathbf{x}_j ,\\
    \mathbf{n}_1 &= (\mathbf{x}_i - \mathbf{x}_j) \times \mathbf{b} ,\\
    \mathbf{n}_2 &= \mathbf{b} \times (\mathbf{x}_l - \mathbf{x}_k).
\end{align}
The Jacobian elements are 
\begin{align}
    \frac{\partial \phi_{ijkl}}{\partial \mathbf{x}_i} &= \frac{\|\mathbf{b}\|}{\|\mathbf{n}_1\|^2} \mathbf{n}_1 ,\\
    \frac{\partial \phi_{ijkl}}{\partial \mathbf{x}_l} &= -\frac{\|\mathbf{b}\|}{\|\mathbf{n}_2\|^2} \mathbf{n}_2 ,\\
    \frac{\partial \phi_{ijkl}}{\partial \mathbf{x}_j} &= \left(\frac{\|\mathbf{x}_k - \mathbf{x}_j\| - \|\mathbf{x}_i - \mathbf{x}_j\| \cos\theta_{ijk}}{\|\mathbf{x}_k - \mathbf{x}_j\| \sin^2\theta_{ijk}}\right) \frac{\partial \phi_{ijkl}}{\partial \mathbf{x}_i} \\&\quad - \left(\frac{\|\mathbf{b}\|}{\|\mathbf{n}_2\|^2}\right) \mathbf{n}_2 ,\\
    \frac{\partial \phi_{ijkl}}{\partial \mathbf{x}_k} &= -\frac{\partial \phi_{ijkl}}{\partial \mathbf{x}_i} - \frac{\partial \phi_{ijkl}}{\partial \mathbf{x}_j} - \frac{\partial \phi_{ijkl}}{\partial \mathbf{x}_l}.
\end{align}

\subsection{Optimal Transport for Molecular Conformations}
\label{app:OT}

The 2-Wasserstein distance between two probability measures $\mu_0, \mu_1$ on $\mathbb{R}^m$ is 
\begin{equation}
    W_2^2(\mu_0, \mu_1) = \inf_{\pi \in \Pi(\mu_0, \mu_1)} \int_{\mathbb{R}^m \times \mathbb{R}^m} \|\mathbf{z}_0 - \mathbf{z}_1\|^2 \, d\pi(\mathbf{z}_0, \mathbf{z}_1).
\end{equation}
The Sinkhorn algorithm converges exponentially to the optimal transport plan 
\begin{equation}
    \|\pi^{(k)} - \pi^*\|_1 \leq C e^{-\rho k},
\end{equation}
where $\rho = \min\left(\frac{\epsilon}{\max_{i,j} C_{ij}}, \frac{1}{\log(n)}\right)$ and $C$ is a problem-dependent constant.

The entropic regularization parameter $\epsilon$ controls the trade-off between optimization accuracy and computational speed 
\begin{itemize}
    \item $\epsilon \to 0$  Recovers exact optimal transport (slow convergence),
    \item $\epsilon \to \infty$  Converges to product measure $\mu_0 \otimes \mu_1$ (fast but inaccurate),
    \item Optimal choice  $\epsilon \approx \sigma^2$ where $\sigma$ is the typical distance scale in the data.
\end{itemize}

\subsection{Flow Matching}
\label{app:FM}

The relationship between our velocity field and the probability density evolution is given by the continuity equation 
\begin{equation}
    \frac{\partial p_t}{\partial t} + \nabla \cdot (p_t v_t) = 0.
\end{equation}

The probability flow ODE that generates samples following $p_t$ is 
\begin{equation}
    \frac{d\mathbf{X}}{dt} = v_t(\mathbf{X}) - \frac{1}{2}\frac{\nabla p_t(\mathbf{X})}{p_t(\mathbf{X})}.
\end{equation}
The flow matching objective 
\begin{equation}
    \mathcal{L}_{FM} = \mathbb{E}_{t,\mathbf{x}_0,\mathbf{x}_1} \left[ \|v_\theta(\mathbf{x}_t, t) - (\mathbf{x}_1 - \mathbf{x}_0)\|^2 \right],
\end{equation}
has the unique global minimum at $v_\theta^*(\mathbf{x}_t, t) = \mathbb{E}[\mathbf{x}_1 - \mathbf{x}_0 | \mathbf{x}_t]$, which coincides with the optimal transport velocity field.



\begin{figure*}[htp]
    \centering
    \includegraphics[width=1\linewidth]{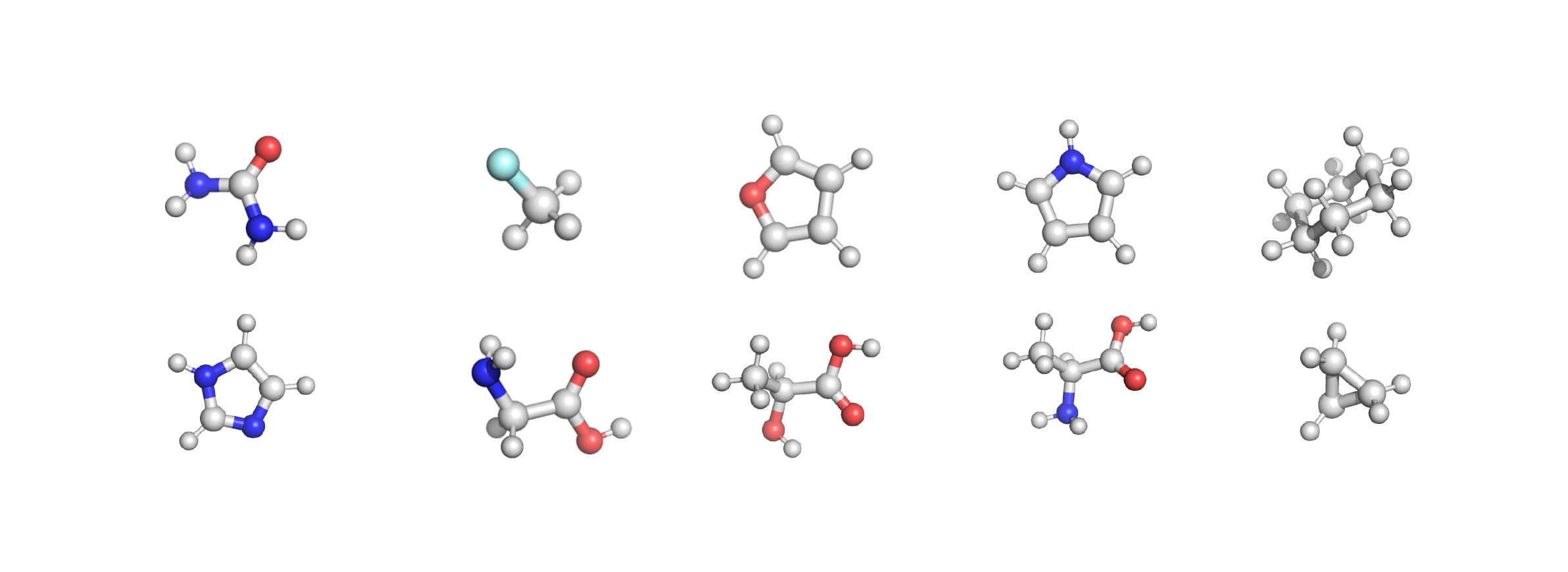}
    \vspace{-1.5cm}
    \caption{Representative samples generated by our model on GEOM-QM9. GO-Flow successfully reconstructs precise local geometries required for quantum chemical calculations.}
    \label{fig:geom-qm9}
\end{figure*}

\section{Implementation Details}
\label{app:details}

\subsection{Datasets} \label{apdx:subsec_data}
We used GEOM-QM9 (QM9) and GEOM-Drugs (Drugs) \cite{axelrod2022geom} for analysis and comparison between molecular conformer generation models. We obtained the raw data, the pre-processed data, and the data split at \url{https://github.com/DeepGraphLearning/ConfGF}. Each dataset comprises 40,000 molecules for the training set and 5,000 molecules for the validation set, with each molecule containing 5 conformers following the protocol of \cite{shi2021learning}. For the test set, we selected 200 molecules for each dataset, resulting in 22,408 and 14,324 conformers existing in QM9 and Drugs, respectively.

\subsection{Input Featurization}
\label{app:input}

Atomic features (or Node Features) are computed using RDKit \citep{landrum2013rdkit} features as described in \autoref{tab:atomic}. For computing edge features and edge index, we use a combination of global (radius based edges) and local (molecular graph edges) similar to \citep{jing2022torsional}.

\subsection{Evaluation Metrics}
\label{app:metrics}
To measure the accuracy and diversity of the generated conformer set $\mathcal{C}$, we adopted metrics proposed by \cite{ganea2021geomol}. The metrics are based on root-mean-square deviation (RMSD), which is a normalized Frobenius norm between two atomic coordinate matrices aligned using the Kabsch algorithm \cite{kabsch1976solution}. Given the ground truth conformer set $\mathcal{C}^*$ and the generated sample set $\mathcal{C}$, four metrics that follow precision and recall are defined as:
\begin{align}
    \text{COV-R} \ (\text{Recall}) &= \frac{1}{\vert \mathcal{C}^* \vert} \vert
    \{C^*\in \mathcal{C}^* \vert \text{RMSD}(C^*, C) \leq \delta,  C \in \mathcal{C} \} \vert, \label{eq:covr} \\
    \text{MAT-R} \ (\text{Recall}) &= \frac{1}{\vert \mathcal{C}^* \vert}
    \sum\limits_{C^* \in \mathcal{C}^*}
    \min\limits_{C \in \mathcal{C}} \text{RMSD}(C^*, C), \label{eq:matr}\\
    \text{COV-P} \ (\text{Precision}) &= \frac{1}{\vert \mathcal{C} \vert} \vert
    \{C\in \mathcal{C} \vert \text{RMSD}(C, C^*) \leq \delta,  C^* \in \mathcal{C^*} \} \vert, \label{eq:covp} \\
    \text{MAT-P} \ (\text{Precision}) &= \frac{1}{\vert \mathcal{C} \vert}
    \sum\limits_{C \in \mathcal{C}}
    \min\limits_{C^* \in \mathcal{C^*}} \text{RMSD}(C, C^*), \label{eq:matp}
\end{align}
where COV and MAT are coverage metric and matching metric \cite{xu2021learning}, respectively. COV quantifies the proportion of one set covered by another, with ``covered'' indicating RMSD values are within a threshold $\delta$. MAT measures the average of RMSD values of one conformer set with its closest conformer in another set. If $\mathcal{C}$ and $\mathcal{C}^*$ are exchanged in Eqs. (\ref{eq:covr}, \ref{eq:matr}), then metrics become COV-P (Precision) and MAT-P (Precision). The recall metric is focused on the diversity, while the precision metric measures the quality. The threshold $\delta$ is set to $0.5\text{\r{A}}$ for QM9 and $1.25\text{\r{A}}$ for Drugs. For each molecule, we generated conformers $C$ that are twice the size of the ground truth conformers $C^*$.

\begin{table}[htb]
\centering
\caption{Hyperparameters of GO-Flow.}
\label{tab:hyper}
\begin{tabular}{lc}
\toprule
Hyperparameter & Value \\
\midrule
Hidden dimension & 128 \\
Number of layers & 8 \\
Batch size & 64 \\
Learning rate (Stage 1) & $10^{-3}$ \\
Learning rate (Stage 2) & $10^{-4}$ \\
Learning rate (Stage 3) & $10^{-5}$ \\
Noise scale (translation) & 0.5 \\
Noise scale (rotation) & 0.3 \\
Noise scale (conformation) & 0.1 \\
Sinkhorn iterations & 100 \\
Entropy regularization $\epsilon$ & 0.1 \\
Training iter. & 700k \\
\bottomrule
\end{tabular}
\end{table}

\subsection{Hyperparameters} \label{app:hyper}

We used eight NVIDIA A6000 (48G) GPUs for the training and generation tasks. We reported hyperparameters of GO-Flow in Table \ref{tab:hyper}.

\section{Visualization on Small Molecules (GEOM-QM9)}
\label{app:visualization_qm9}
Appendix Figure \ref{fig:geom-qm9} displays representative 3D conformations generated by GO-Flow on the GEOM-QM9 dataset. As observed in the samples, our model successfully reconstructs the precise local geometries required for quantum chemical calculations. The generated molecules exhibit valid bond lengths and angles, devoid of the disconnected components or distorted rings often seen in baseline methods. This visual evidence supports our quantitative findings (Table~\ref{tab:rmsd_qm9}) that GO-Flow achieves sub-Angstrom matching accuracy by respecting the intrinsic constraints of the molecular manifold.

\newpage

%% file: example_paper.bib
@article{liu2023generative,
  title={Generative diffusion models on graphs: Methods and applications},
  author={Liu, Chengyi and Fan, Wenqi and Liu, Yunqing and Li, Jiatong and Li, Hang and Liu, Hui and Tang, Jiliang and Li, Qing},
  journal={arXiv preprint arXiv:2302.02591},
  year={2023}
}

@article{he2025graph,
  title={Graph Defense Diffusion Model},
  author={He, Xin and Fan, Wenqi and Wang, Yili and Liu, Chengyi and Miao, Rui and Juan, Xin and Wang, Xin},
  journal={arXiv preprint arXiv:2501.11568},
  year={2025}
}

@article{liu2025score,
  title={Score-based generative diffusion models for social recommendations},
  author={Liu, Chengyi and Zhang, Jiahao and Wang, Shijie and Fan, Wenqi and Li, Qing},
  journal={IEEE Transactions on Knowledge and Data Engineering},
  year={2025},
  publisher={IEEE}
}

@article{fan2025computational,
  title={Computational protein science in the era of large language models (llms)},
  author={Fan, Wenqi and Zhou, Yi and Wang, Shijie and Yan, Yuyao and Liu, Hui and Zhao, Qian and Song, Le and Li, Qing},
  journal={arXiv preprint arXiv:2501.10282},
  year={2025}
}

@article{sadybekov2023computational,
  title={Computational approaches streamlining drug discovery},
  author={Sadybekov, Anastasiia V and Katritch, Vsevolod},
  journal={Nature},
  volume={616},
  number={7958},
  pages={673--685},
  year={2023},
  publisher={Nature Publishing Group UK London}
}

@article{dara2022machine,
  title={Machine learning in drug discovery: a review},
  author={Dara, Suresh and Dhamercherla, Swetha and Jadav, Surender Singh and Babu, CH Madhu and Ahsan, Mohamed Jawed},
  journal={Artificial intelligence review},
  volume={55},
  number={3},
  pages={1947--1999},
  year={2022},
  publisher={Springer}
}

@article{vincent2022phenotypic,
  title={Phenotypic drug discovery: recent successes, lessons learned and new directions},
  author={Vincent, Fabien and Nueda, Arsenio and Lee, Jonathan and Schenone, Monica and Prunotto, Marco and Mercola, Mark},
  journal={Nature Reviews Drug Discovery},
  volume={21},
  number={12},
  pages={899--914},
  year={2022},
  publisher={Nature Publishing Group UK London}
}

@inproceedings{
xu2022geodiff,
title={GeoDiff: A Geometric Diffusion Model for Molecular Conformation Generation},
author={Minkai Xu and Lantao Yu and Yang Song and Chence Shi and Stefano Ermon and Jian Tang},
booktitle={International Conference on Learning Representations},
year={2022},
url={https://openreview.net/forum?id=PzcvxEMzvQC}
}

@article{ganea2021geomol,
  title={Geomol: Torsional geometric generation of molecular 3d conformer ensembles},
  author={Ganea, Octavian and Pattanaik, Lagnajit and Coley, Connor and Barzilay, Regina and Jensen, Klavs and Green, William and Jaakkola, Tommi},
  journal={Advances in Neural Information Processing Systems},
  volume={34},
  pages={13757--13769},
  year={2021}
}

@inproceedings{shi2021learning,
  title={Learning gradient fields for molecular conformation generation},
  author={Shi, Chence and Luo, Shitong and Xu, Minkai and Tang, Jian},
  booktitle={International conference on machine learning},
  pages={9558--9568},
  year={2021},
  organization={PMLR}
}

@article{luo2021predicting,
  title={Predicting molecular conformation via dynamic graph score matching},
  author={Luo, Shitong and Shi, Chence and Xu, Minkai and Tang, Jian},
  journal={Advances in neural information processing systems},
  volume={34},
  pages={19784--19795},
  year={2021}
}

@article{jing2022torsional,
  title={Torsional diffusion for molecular conformer generation},
  author={Jing, Bowen and Corso, Gabriele and Chang, Jeffrey and Barzilay, Regina and Jaakkola, Tommi},
  journal={Advances in neural information processing systems},
  volume={35},
  pages={24240--24253},
  year={2022}
}

@article{hassan2024flow,
  title={Et-flow: Equivariant flow-matching for molecular conformer generation},
  author={Hassan, Majdi and Shenoy, Nikhil and Lee, Jungyoon and St{\"a}rk, Hannes and Thaler, Stephan and Beaini, Dominique},
  journal={Advances in Neural Information Processing Systems},
  volume={37},
  pages={128798--128824},
  year={2024}
}

@article{axelrod2022geom,
  title={GEOM, energy-annotated molecular conformations for property prediction and molecular generation},
  author={Axelrod, Simon and Gomez-Bombarelli, Rafael},
  journal={Scientific Data},
  volume={9},
  number={1},
  pages={185},
  year={2022},
  publisher={Nature Publishing Group UK London}
}

@article{park2024equivariant,
  title={Equivariant blurring diffusion for hierarchical molecular conformer generation},
  author={Park, Jiwoong and Shen, Yang},
  journal={Advances in Neural Information Processing Systems},
  volume={37},
  pages={131645--131675},
  year={2024}
}

@article{mansimov2019molecular,
  title={Molecular geometry prediction using a deep generative graph neural network},
  author={Mansimov, Elman and Mahmood, Omar and Kang, Seokho and Cho, Kyunghyun},
  journal={Scientific reports},
  volume={9},
  number={1},
  pages={20381},
  year={2019},
  publisher={Nature Publishing Group UK London}
}

@InProceedings{simm2019generative,
  title = 	 {A Generative Model for Molecular Distance Geometry},
  author =       {Simm, Gregor and Hernandez-Lobato, Jose Miguel},
  booktitle = 	 {Proceedings of the 37th International Conference on Machine Learning},
  pages = 	 {8949--8958},
  year = 	 {2020},
  editor = 	 {III, Hal Daumé and Singh, Aarti},
  volume = 	 {119},
  series = 	 {Proceedings of Machine Learning Research},
  month = 	 {13--18 Jul},
  publisher =    {PMLR},
  url = 	 {https://proceedings.mlr.press/v119/simm20a.html}
}

@inproceedings{
xu2021learning,
title={Learning Neural Generative Dynamics for Molecular Conformation Generation},
author={Minkai Xu and Shitong Luo and Yoshua Bengio and Jian Peng and Jian Tang},
booktitle={International Conference on Learning Representations},
year={2021},
url={https://openreview.net/forum?id=pAbm1qfheGk}
}

@inproceedings{xu2021end,
  title={An end-to-end framework for molecular conformation generation via bilevel programming},
  author={Xu, Minkai and Wang, Wujie and Luo, Shitong and Shi, Chence and Bengio, Yoshua and Gomez-Bombarelli, Rafael and Tang, Jian},
  booktitle={International conference on machine learning},
  pages={11537--11547},
  year={2021},
  organization={PMLR}
}

@inproceedings{
liu2026enhancing,
title={Enhancing Molecular Property Predictions by Learning from Bond Modelling and Interactions},
author={Liu, Yunqing and Zhou, Yi and Fan, Wenqi},
booktitle={The Fourteenth International Conference on Learning Representations},
year={2026},
url={https://openreview.net/forum?id=S4bJQ4p9hx}
}

@article{weng2021late,
  title={Late-stage photoredox C--H amidation of N-unprotected indole derivatives: Access to n-(indol-2-yl) amides},
  author={Weng, Yue and Ding, Bo and Liu, Yunqing and Song, Chunlan and Chan, Lo-Ying and Chiang, Chien-Wei},
  journal={Organic Letters},
  volume={23},
  number={7},
  pages={2710--2714},
  year={2021},
  publisher={ACS Publications}
}

@article{ding2019selective,
  title={Selective photoredox trifluoromethylation of tryptophan-containing peptides},
  author={Ding, Bo and Weng, Yue and Liu, Yunqing and Song, Chunlan and Yin, Le and Yuan, Jiafan and Ren, Yanrui and Lei, Aiwen and Chiang, Chien-Wei},
  journal={European Journal of Organic Chemistry},
  volume={2019},
  number={46},
  pages={7596--7605},
  year={2019},
  publisher={Wiley Online Library}
}

@inproceedings{wang2024swallowing,
    title={Swallowing the Bitter Pill: Simplified Scalable Conformer Generation},
    author={Wang, Yuyang and Elhag, Ahmed A. and Jaitly, Navdeep and Susskind, Joshua M. and Bautista, Miguel {\'A}ngel},
    year={2024},
    booktitle={Forty-first International Conference on Machine Learning},
}

@article{landrum2013rdkit,
  title={RDKit: A software suite for cheminformatics, computational chemistry, and predictive modeling},
  author={Landrum, Greg and others},
  journal={Greg Landrum},
  volume={8},
  number={31.10},
  pages={5281},
  year={2013}
}

@article{hawkins2010conformer,
  title={Conformer generation with OMEGA: algorithm and validation using high quality structures from the Protein Databank and Cambridge Structural Database},
  author={Hawkins, Paul CD and Skillman, A Geoffrey and Warren, Gregory L and Ellingson, Benjamin A and Stahl, Matthew T},
  journal={Journal of chemical information and modeling},
  volume={50},
  number={4},
  pages={572--584},
  year={2010},
  publisher={ACS Publications}
}

@article{cao2025efficient,
  title={Efficient molecular conformer generation with so (3)-averaged flow matching and reflow},
  author={Cao, Zhonglin and Geiger, Mario and Costa, Allan Dos Santos and Reidenbach, Danny and Kreis, Karsten and Geffner, Tomas and Pellegrini, Franco and Zhou, Guoqing and Kucukbenli, Emine},
  journal={arXiv preprint arXiv:2507.09785},
  year={2025}
}

@article{hollingsworth2018molecular,
  title={Molecular dynamics simulation for all},
  author={Hollingsworth, Scott A and Dror, Ron O},
  journal={Neuron},
  volume={99},
  number={6},
  pages={1129--1143},
  year={2018},
  publisher={Elsevier}
}

@incollection{singh2022molecular,
  title={Molecular docking and molecular dynamics simulation},
  author={Singh, Sakshi and Baker, Qanita Bani and Singh, Dev Bukhsh},
  booktitle={Bioinformatics},
  pages={291--304},
  year={2022},
  publisher={Elsevier}
}

@article{ho2020denoising,
  title={Denoising diffusion probabilistic models},
  author={Ho, Jonathan and Jain, Ajay and Abbeel, Pieter},
  journal={Advances in neural information processing systems},
  volume={33},
  pages={6840--6851},
  year={2020}
}

@inproceedings{
lipman2023flow,
title={Flow Matching for Generative Modeling},
author={Yaron Lipman and Ricky T. Q. Chen and Heli Ben-Hamu and Maximilian Nickel and Matthew Le},
booktitle={The Eleventh International Conference on Learning Representations },
year={2023},
url={https://openreview.net/forum?id=PqvMRDCJT9t}
}

@article{kabsch1976solution,
  title={A solution for the best rotation to relate two sets of vectors},
  author={Kabsch, Wolfgang},
  journal={Foundations of Crystallography},
  volume={32},
  number={5},
  pages={922--923},
  year={1976},
  publisher={International Union of Crystallography}
}

@article{smith2020psi4,
  title={PSI4 1.4: Open-source software for high-throughput quantum chemistry},
  author={Smith, Daniel GA and Burns, Lori A and Simmonett, Andrew C and Parrish, Robert M and Schieber, Matthew C and Galvelis, Raimondas and Kraus, Peter and Kruse, Holger and Di Remigio, Roberto and Alenaizan, Asem and others},
  journal={The Journal of chemical physics},
  volume={152},
  number={18},
  year={2020},
  publisher={AIP Publishing}
}

@InProceedings{pmlr-v235-geist24a,
  title = 	 {Learning with 3{D} rotations, a hitchhiker’s guide to {SO}(3)},
  author =       {Geist, Andreas Ren\'{e} and Frey, Jonas and Zhobro, Mikel and Levina, Anna and Martius, Georg},
  booktitle = 	 {Proceedings of the 41st International Conference on Machine Learning},
  pages = 	 {15331--15350},
  year = 	 {2024},
  editor = 	 {Salakhutdinov, Ruslan and Kolter, Zico and Heller, Katherine and Weller, Adrian and Oliver, Nuria and Scarlett, Jonathan and Berkenkamp, Felix},
  volume = 	 {235},
  series = 	 {Proceedings of Machine Learning Research},
  month = 	 {21--27 Jul},
  publisher =    {PMLR},
  url = 	 {https://proceedings.mlr.press/v235/geist24a.html}
}

@article{hemingway2018perspectives,
  title={Perspectives on Euler angle singularities, gimbal lock, and the orthogonality of applied forces and applied moments},
  author={Hemingway, Evan G and O’Reilly, Oliver M},
  journal={Multibody system dynamics},
  volume={44},
  number={1},
  pages={31--56},
  year={2018},
  publisher={Springer}
}

@article{buss2001spherical,
  title={Spherical averages and applications to spherical splines and interpolation},
  author={Buss, Samuel R and Fillmore, Jay P},
  journal={ACM Transactions on Graphics (TOG)},
  volume={20},
  number={2},
  pages={95--126},
  year={2001},
  publisher={ACM New York, NY, USA}
}

@inproceedings{liu-etal-2025-glprotein,
    title = "{GLP}rotein: Global-and-Local Structure Aware Protein Representation Learning",
    author = "Liu, Yunqing  and
      Fan, Wenqi  and
      Wei, Xiaoyong  and
      Qing, Li",
    booktitle = "Findings of the Association for Computational Linguistics: EMNLP 2025",
    month = nov,
    year = "2025",
    address = "Suzhou, China",
    publisher = "Association for Computational Linguistics",
}

@article{zhao2024recommender,
  title={Recommender systems in the era of large language models (llms)},
  author={Zhao, Zihuai and Fan, Wenqi and Li, Jiatong and Liu, Yunqing and Mei, Xiaowei and Wang, Yiqi and Wen, Zhen and Wang, Fei and Zhao, Xiangyu and Tang, Jiliang and others},
  journal={IEEE Transactions on Knowledge and Data Engineering},
  volume={36},
  number={11},
  pages={6889--6907},
  year={2024},
  publisher={IEEE}
}

@article{li2024empowering,
  title={Empowering molecule discovery for molecule-caption translation with large language models: A chatgpt perspective},
  author={Li, Jiatong and Liu, Yunqing and Fan, Wenqi and Wei, Xiao-Yong and Liu, Hui and Tang, Jiliang and Li, Qing},
  journal={IEEE transactions on knowledge and data engineering},
  volume={36},
  number={11},
  pages={6071--6083},
  year={2024},
  publisher={IEEE}
}

@inproceedings{hu2023improving,
  title={Improving user controlled table-to-text generation robustness},
  author={Hu, Hanxu and Liu, Yunqing and Yu, Zhongyi and Perez-Beltrachini, Laura},
  booktitle={Findings of the Association for Computational Linguistics: EACL 2023},
  pages={2317--2324},
  year={2023}
}

@article{li2024tomg,
  title={Tomg-bench: Evaluating llms on text-based open molecule generation},
  author={Li, Jiatong and Li, Junxian and Liu, Yunqing and Zhou, Dongzhan and Li, Qing}
}

@article{li2026molreflect,
  title={Molreflect: Towards in-context fine-grained alignments between molecules and texts},
  author={Li, Jiatong and Liu, Yunqing and Liu, Wei and Lei, Jingdi and Zhang, Di and Fan, Wenqi and Zhou, Dongzhan and Li, Yuqiang and Li, Qing},
  journal={IEEE Transactions on Knowledge and Data Engineering},
  year={2026},
  publisher={IEEE}
}

@article{li2024speak,
  title={Speak-to-Structure: Evaluating LLMs in Open-domain Natural Language-Driven Molecule Generation},
  author={Li, Jiatong and Li, Junxian and Wang, Weida and Liu, Yunqing and Zheng, Changmeng and Zhou, Dongzhan and Wei, Xiao-yong and Li, Qing},
  journal={arXiv preprint arXiv:2412.14642},
  year={2024}
}
